\documentclass[10pt,twocolumn,twoside]{IEEEtran}
\usepackage{cite}
\usepackage{amsmath,amssymb,amsfonts}
\usepackage{algorithmic}
\usepackage{graphicx}
\usepackage{textcomp}
\usepackage{array}
\usepackage{subfig}
\usepackage{textcomp}
\usepackage{stfloats}
\usepackage{url}
\usepackage{verbatim}
\usepackage{balance}
\usepackage{nicefrac}
\usepackage[commandnameprefix=always]{changes}
\usepackage{bm}
\usepackage{amsthm}
\usepackage{amssymb}
\usepackage{mathtools}
\usepackage[colorlinks=true, allcolors=blue]{hyperref}
\usepackage{bbm}
\usepackage{url}
\usepackage{caption}
\usepackage{algorithm}
\usepackage{verbatim}
\usepackage{xcolor}
 
\newtheorem{theorem}{Theorem}
\newtheorem{lemma}{Lemma}

\newtheorem{proposition}{Proposition}
\newtheorem{assumption}{Assumption}
\newtheorem{remark}{Remark}

\def\BibTeX{{\rm B\kern-.05em{\sc i\kern-.025em b}\kern-.08em
    T\kern-.1667em\lower.7ex\hbox{E}\kern-.125emX}}
\begin{document}

\title{Parallel Momentum Methods Under Biased Gradient Estimations}
\author{Ali Beikmohammadi,
Sarit Khirirat, and Sindri Magn\'usson 
\thanks{This work was partially supported by the Swedish Research Council (2020-03607), Digital Futures, and  
Sweden's Innovation Agency (Vinnova). 
The computations were enabled by resources provided by the National Academic Infrastructure for Supercomputing in Sweden (NAISS) at Chalmers Centre for Computational Science and Engineering (C3SE) partially funded by the Swedish Research Council through grant agreement no. 2022-06725.}
\thanks{A. Beikmohammadi and S. Magn\'usson are with the Department of Computer and System Science, Stockholm University, 11419 Stockholm, Sweden (e-mail: beikmohammadi@dsv.su.se; sindri.magnusson@dsv.su.se).}
\thanks{S. Khirirat is a postdoctoral fellow at King Abdullah University of Science and Technology (KAUST), Thuwal, Saudi Arabia (e-mail: sarit.khirirat@kaust.edu.sa). He contributed to this paper before joining KAUST.}
}

\maketitle

\begin{abstract}
Parallel stochastic gradient methods are gaining prominence in solving large-scale machine learning problems that involve data distributed across multiple nodes. However, obtaining unbiased stochastic gradients, which have been the focus of most theoretical research, is challenging in many distributed machine learning applications. The gradient estimations easily become biased, for example, when gradients are compressed or clipped, when data is shuffled, and in meta-learning and reinforcement learning. 
In this work, we establish worst-case bounds on  parallel momentum methods under biased gradient estimation on both general non-convex and $\mu$-PL problems. Our analysis covers general distributed optimization problems, and we work out the implications for special cases where gradient estimates are biased, i.e. in meta-learning and when the gradients are compressed or clipped. Our numerical experiments verify our theoretical findings and show faster convergence performance of momentum methods than traditional biased gradient descent.
\end{abstract}


\section{Introduction}
\IEEEPARstart{T}{he} increasing scale of machine learning models in data samples and model parameters can significantly improve classification accuracy \cite{le2011optimization,coates2011analysis}. 
This motivates the development of learning algorithms under the server-worker architecture, where computing nodes collaboratively optimize the parameters of specific learning models.
In particular, if we have $n$ nodes, the goal is to find the learning model parameters $x\in\mathbb{R}^d$ that minimize the average of the loss functions of all $n$ nodes:
\begin{align}\label{eqn:Problem}
	\mathop{\text{minimize}}\limits_{x\in\mathbb{R}^d} \quad f(x) := \frac{1}{n}\sum_{i=1}^n f_i(x),
\end{align}	
where $f_i(x)$ is the loss function based on the data locally stored on node $i$.

To solve large-scale distributed optimization problems,  parallel stochastic gradient descent (SGD) is among the most popular algorithms.
A common framework for implementing  parallel SGD is a parameter server  \cite{li2013parameter}, which comprises a central server and worker nodes. 
At each iteration of  parallel SGD, each worker node computes a stochastic gradient based on its private data. The central server then updates the model parameters $x^k$ according to: 
\begin{align}\label{eqn:BiasedGradient}
x^{k+1} = x^k - \frac{\gamma}{n}\sum_{i=1}^n \tilde \nabla f_i(x^k),
\end{align} 
where $\gamma$ is a positive step-size and $\tilde \nabla f_i(x)$ is the stochastic gradient estimator for $\nabla f_i(x)$ evaluated by worker node $i$. 

Among many efforts to improve the convergence speed and solution accuracy of parallel SGD, parallel momentum methods are known as quite well-established techniques. 
Particularly, in parallel momentum methods, at each iteration, every worker node computes its stochastic gradient $\tilde \nabla f_i(x)$. The central server then updates the gradient estimate $v^k$ using a convex combination of its previous estimate $v^{k-1}$ and the  aggregated gradients $(1/n)\sum_{i=1}^n \tilde \nabla f_i(x^k)$, weighted by the momentum parameter $\beta\in(0,1]$:
\begin{align}
  v^k  &= v^{k-1} + \beta \left( \frac{1}{n}\sum_{i=1}^n \tilde \nabla f_i(x^k) - v^{k-1}\right).\label{eqn:momentum_v^k}
\end{align}
Then, the server updates the model parameters $x^k$ via:
\begin{align}
    x^{k+1}  & = x^k - \gamma v^k. \label{eqn:momentum_x^k} 
\end{align}
Note that momentum methods with $\beta=1$ recover SGD \eqref{eqn:BiasedGradient}.  The superior performance of momentum methods compared to SGD has been shown experimentally in neural network training in~\cite{krizhevsky2017imagenet,sutskever2013importance}.

Multiple works have studied theoretical convergence guarantees of SGD and its variance, including momentum methods. A prevalent assumption within these works has been that the stochastic gradients are unbiased, i.e.  that $\mathbf{E}[ \tilde \nabla f_i(x)]=\nabla f_i(x)$ for $x\in\mathbb{R}^d$, \cite{bottou2012stochastic,bottou2010large,zinkevich2010parallelized,loshchilov2016sgdr,feyzmahdavian2016asynchronous,khirirat2017mini,ge2019step,wang2021convergence,sebbouh2021almost,woodworth2020minibatch,csiba2018importance,xiao2014proximal,johnson2013accelerating}. 
However, gradient estimators exhibit bias in various machine learning applications. 
For instance, when performing random shuffling, without-replacement sampling, or cyclic sampling of gradients, the resulting estimators exhibit bias~\cite{mishchenko2020random,haochen2019random,gurbuzbalaban2021random}. Furthermore, the use of compression operators to enhance communication efficiency can introduce bias in gradient estimation~\cite{alistarh2017qsgd,beznosikov2020biased}. Similarly, clipping operators employed to stabilize the training of deep neural networks introduce bias~\cite{pascanu2013difficulty}. Biased gradient estimators are also commonly observed in adversarial learning with byzantine SGD, as well as in meta-learning and reinforcement learning applications~\cite{alistarh2018byzantine,pillutla2022robust,li2019rsa, finn2017model}.

Already in the early eighties, Ruszczy\'{n}ski and Syski illustrated the superior behavior of momentum methods over SGD under biased gradients~\cite{ruszczynski1983stochastic}. Their work demonstrated that momentum methods are capable of converging to stationary points \cite{ruszczynski1983stochastic}. However, the study of the worst-case bound of this case, alongside its parallel setting, has yet remained underexplored.

\subsection{Contributions}
\noindent We provide worst-case bounds for parallel  momentum methods with general biased gradient models for general non-convex and $\mu$-PL problems. 
We show that momentum methods with biased gradients enjoy the bound similar to SGD with biased gradients in \cite{ajalloeian2020convergence}. 
While \cite{liu2020improved} holds only for centralized settings, our analysis framework applies to the server-worker architectures. 
The analysis does not require the unbiased property of gradient estimators and relies on the inequality in Lemma~\ref{lemma:descent} to derive the worst-case bounds under general and PL non-convex problems. 
We apply our unified theory to momentum methods with three popular biased gradient examples: gradient compression, gradient clipping, and stochastic composite gradients, including meta-learning.  
We finally corroborate theoretical findings with the experiment and demonstrate the superior performance of  parallel momentum methods over  parallel SGD under biased gradient estimation on training neural networks in a distributed setup.

\subsection{Organization}
\noindent  We begin by reviewing related literature on stochastic momentum methods and other variants in Section~\ref{sec:liter}. 
We next present unified convergence results for  parallel stochastic momentum methods with biased gradients for general non-convex and $\mu$-PL problems in Section~\ref{sec:convergence}. 
In Section~\ref{sec:App},
we then show how to apply our unified framework to establish the convergence of  parallel stochastic momentum methods using compressed gradients, clipped gradients, and stochastic composite gradients. 
Finally, we present in Section~\ref{sec:exp} significant convergence improvement from  parallel momentum methods over traditional  parallel SGD for training deep neural network models and then conclude in Section \ref{sec:conclusion}. 

\section{Related Works}\label{sec:liter}
\noindent We review literature closely related to momentum methods and other variants under both unbiased and biased gradient estimation assumptions and contrast them with our work.

\subsection{Unbiased Gradient Estimation}
\noindent Studies on the convergence behaviors of SGD with unbiased gradient estimation have been well-established \cite{bottou2012stochastic,bottou2010large,zinkevich2010parallelized,loshchilov2016sgdr,feyzmahdavian2016asynchronous,khirirat2017mini,ge2019step,wang2021convergence,sebbouh2021almost,lei2019stochastic,khaled2022better}. 
Specifically, SGD with a constant step-size was shown to converge close to the optimal solution up to a residual error \cite{needell2014stochastic,bertsekas2011incremental}. 
To improve the solution accuracy of SGD, popular approaches include  mini-batching \cite{woodworth2020minibatch,csiba2018importance}, variance reduction \cite{xiao2014proximal,johnson2013accelerating}, and momentum techniques \cite{polyak1964some,nesterov1983method,gupal1972stochastic}. In this paper, we focus on the last approach, which leads to SGD using momentum, known as the stochastic momentum method. 
The convergence of stochastic momentum methods was originally analyzed by \cite{gupal1972stochastic} and later has been refined and shown to converge theoretically as fast as SGD under the same setting \cite{yan2018unified,yu2019linear,liu2020improved}. 
For instance, in \cite{liu2020improved}, stochastic momentum methods achieve the convergence bound similar to SGD at $\mathcal{O}({1}/(k\gamma) + {\gamma\sigma^2}/{(1-\beta)})$ and $\mathcal{O}((1-\gamma\mu)^k + \gamma\sigma^2)$ for non-convex and strongly convex problems under standard assumptions, respectively, where $\sigma^2= \max_x \mathbf{E}\| \tilde \nabla f(x) - \nabla f(x)\|^2$.
Additionally, methods using other momentum variants, including Polyak's Heavy-ball momentum \cite{polyak1964some}, Nesterov's momentum \cite{nesterov1983method}, and other momentum schemes such as Quasi-Hyperbolic momentum \cite{ma2018quasi}, and PID control-based momentum \cite{wang2020pid}, have been studied under various and more general problem setups, e.g. in \cite{sebbouh2021almost, ghadimi2015global,loizou2017linearly,assran2020convergence,xin2019distributed}.
For example, in \cite{ghadimi2015global}, for solving strongly convex quadratic problems, Heavy-ball momentum and Nesterov's momentum methods converge at the $(\sqrt{L/\mu}-1)/(\sqrt{L/\mu}+1)$ and $(\sqrt{L/\mu}-1)/(\sqrt{L/\mu})$ rate, respectively. 
Unlike these works, our paper shows worst-case bounds for stochastic momentum methods under general biased gradient for distributed optimization, which is both a more realistic and relaxed setup. 
Note that, in this paper, although we restrict ourselves among momentum variants to stochastic momentum methods, our result can be applied to the methods with other momentum variants, given specific choices of hyper-parameters shown by \cite{garrigos2023handbook}.

\subsection{Biased Gradient Estimation - Specific Cases}
\noindent Although many works analyzed SGD and stochastic momentum methods with unbiased estimators, there are limited studies of these methods under biased gradients. 
Even among them, the convergence analysis has been mainly restricted to specific types of bias caused by gradient estimators, e.g. random shuffling (which in theory covers without-replacement sampling and cyclic sampling of gradients) \cite{gurbuzbalaban2021random,mishchenko2020random,haochen2019random,tran2021smg}, compression \cite{karimireddy2019error,beznosikov2020biased}, and clipping \cite{zhang2019gradient,mai2021stability,zhang2020improved}. 
Of these, only~\cite{tran2021smg} considers momentum methods; the rest are restricted to SGD. 
In this paper, rather than analyzing stochastic momentum methods under specific types of biased gradients, we unify the analysis framework by relying on the general biased gradient model and considering the server-worker setting.

\subsection{Biased Gradient Estimation - General Cases}
\noindent Despite works done on particular choices of biased gradients, to the best of our knowledge, very few works unified the analysis framework for SGD and momentum methods with biased gradient estimations on a generic form, regardless of their roots \cite{bertsekas2000gradient, ajalloeian2020convergence, ruszczynski1983stochastic}.
On the one hand, 
non-asymptotic convergence of SGD was shown when biased gradients exploited variance-reduction techniques ~\cite{driggs2022biased} and the composite problem structure~\cite{hu2020biased}. On the other hand, SGD with generic biased gradients was first shown to diverge or converge to a finite value by \cite{bertsekas2000gradient}. Recent non-asymptotic convergence to a neighborhood of the solution due to the gradient bias was proved for centralized SGD by \cite{demidovich2024guide,ajalloeian2020convergence}. Stochastic momentum methods with general biased gradients were only studied by \cite{ruszczynski1983stochastic}, demonstrating eventual convergence toward a stationary point of the objective function. However, their result only provides local convergence, without an explicit convergence rate, and is valid when the initialization is sufficiently close to a stationary point.
In contrast to the aforementioned work, our paper focuses on 
the analysis for stochastic momentum methods under biased gradient models under the server-worker setup, for both general non-convex problems and $\mu$-PL problems. It is worth mentioning that our results cover SGD as well by setting $\beta=1$. 

\section{Convergence of Parallel Momentum Methods with Biased Gradients}\label{sec:convergence}
\noindent We present unified convergence for the parallel momentum methods with biased gradients in Eq.~\eqref{eqn:momentum_x^k} and \eqref{eqn:momentum_v^k} on general non-convex and $\mu$-PL problems in Eq.~\eqref{eqn:Problem}. 
We begin by introducing the following standard assumptions on objective functions used throughout this paper.

\begin{assumption}[Smoothness and lower boundedness]\label{assum:smooth_bounded}
	The objective function $f(x)$ is bounded from below by $f^\star= \inf_{x\in\mathbb{R}^d} f(x)>-\infty$, and has $L$-Lipschitz continuous gradient, i.e. $\Vert \nabla f(x) - \nabla f(y) \Vert \leq L \Vert x -y \Vert$ for all $x,y\in\mathbb{R}^d$.
\end{assumption}	

Assumption \ref{assum:smooth_bounded} implies the following inequality: 
\begin{align}\label{eqn:smoothness_ineq}
	f(y) \leq f(x) + \langle \nabla f(x), y-x \rangle + \frac{L}{2}\Vert y-x \Vert^2, \quad \forall x,y\in\mathbb{R}^d.
\end{align}	
Next, we introduce the Polyak-Łojasiewicz (PL) condition on the objective function $f(x)$.
\begin{assumption}\label{assum:PL}
	The  objective function $f(x)$ satisfies the  Polyak-Łojasiewicz (PL) condition with $\mu>0$ if for all $x\in\mathbb{R}^d$
\begin{align}\label{eqn:PL}
	\| \nabla f(x) \|^2 \geq 2\mu [ f(x) - f^\star],
\end{align}	
where $f^\star= \inf_{x\in\mathbb{R}^d} f(x)$.
\end{assumption}	

Assumptions \ref{assum:smooth_bounded} and  \ref{assum:PL} are commonly used to derive the convergence of optimization methods~\cite{karimi2016linear,foster2018uniform,lei2019stochastic},
since they cover many distributed learning problems of interest. 
In particular, Assumption~\ref{assum:smooth_bounded} implies the smoothness of the function $f(x)$, while
Assumption~\ref{assum:PL} is satisfied if $f(x)$ is $\mu$-strongly convex \cite{karimi2016linear}. 
Problems satisfying Assumption \ref{assum:smooth_bounded} include classification with a non-convex regularization (i.e. $\sum_{j=1}^d {x_j^2}/{(1+x_j^2)}$) and training  deep neural networks~\cite{herrera2020estimating}.  
Problems satisfying Assumptions \ref{assum:smooth_bounded} and \ref{assum:PL} include $\ell_2$-regularized empirical risk minimization in distributed settings, i.e. Problem~\eqref{eqn:Problem} with $$f_i(x)= \frac{1}{m}\sum_{j=1}^m \phi( \langle a_{i,j},x \rangle,b_{i,j}) + \frac{\lambda}{2}\| x \|^2,$$ where
$\phi:\mathbb{R}\times\mathbb{R}\rightarrow\mathbb{R}$ is a differentiable function, ${(a_{i,1},b_{i,1}),\ldots,(a_{i,m},b_{i,m})}$ is the set of $m$ data points locally known by worker node $i$, $a_{i,j}\in\mathbb{R}^d$ is the $j^{\rm th}$-data point at worker node $i$ with its associated label $b_{i,j}\in\{-1,1\}$, and $\lambda>0$ is an $\ell_2$-regularization parameter. 
This problem formulation covers  $\ell_2$-regularized linear least-squares problems when $\phi(x,y)=(1/2)(x-y)^2$, and $\ell_2$-regularized logistic regression problems when $\phi(x,y) = \log(1+\exp(-x \cdot y))$.

\textbf{\textit{Key descent lemma for momentum methods.}} To facilitate our analysis, we consider the equivalent update for  parallel momentum methods in Eq.  \eqref{eqn:momentum_x^k} and \eqref{eqn:momentum_v^k} below:
\begin{align}
	x^{k+1} &= x^k - \gamma v^k \quad \text{and} \label{eqn:momentum_equivalent_x_k} \\
 v^k &= v^{k-1} + \beta(\nabla f(x^k)+\eta^k - v^{k-1}), \label{eqn:momentum_equivalent_v_k}
\end{align}	
where 
\begin{align}\label{eqn:generic_eta_k}
   \eta^k=(1/n)\sum_{i=1}^n \tilde \nabla f_i(x^k) - \nabla f(x^k).
\end{align}

To establish the convergence for the momentum method in Eq. \eqref{eqn:momentum_equivalent_x_k} and \eqref{eqn:momentum_equivalent_v_k} under biased gradient estimation, we use the following lemma that is a key to our analysis.

\begin{lemma}\label{lemma:descent}
	Consider the momentum methods in 
 Eq. \eqref{eqn:momentum_equivalent_x_k} and \eqref{eqn:momentum_equivalent_v_k}  for Problem \eqref{eqn:Problem} where Assumption~\ref{assum:smooth_bounded} holds. Let $\phi^k = f(x^k) - f^\star + A \| \nabla f(x^k) - v^{k-1}\|^2$ for $A>0$ and $\eta^{k}$  in~\eqref{eqn:generic_eta_k}. Then, 
	\begin{align}
		\phi^{k+1} 
	&\leq f(x^k)-f^\star - \frac{\gamma}{2}\| \nabla f(x^k) \|^2 + B_1 \| \nabla f(x^k)-v^{k-1} \|^2 \nonumber \\
	&\hspace{0.5cm}- B_2\| x^{k+1} - x^k \|^2 +B_3 \| \eta^k\|^2,\label{eqn:descent}
	\end{align}	
where $B_1 = \gamma \frac{1-\beta}{2} + A\left(1-\frac{\beta}{2}\right)$, $B_2 = \frac{1}{2\gamma} - \frac{L}{2}  - A \frac{(\beta+2)L^2}{\beta}$ and $B_3=\frac{\gamma\beta}{2} + A\beta\left(1+\frac{\beta}{2}\right)$.
\end{lemma}	
\begin{proof}
        See Section \ref{app:lemma:descent}.
\end{proof}	

From Lemma \ref{lemma:descent}, we establish the inequality for deriving convergence theorems for momentum methods. Finally, we introduce the assumption on the upper bound for $\mathbf{E}\| \eta^k \|^2$. 
\begin{assumption}[Affine variance]\label{assum:noise}
For fixed constants $B,C \geq 0$, the variance of stochastic gradients at any iteration $k$ satisfies
 \begin{align*}
     \mathbf{E}\| \eta^k \|^2 \leq B \mathbf{E}\| \nabla f(x^k)\|^2 + C.
 \end{align*}
\end{assumption}

Like the assumption on the biased gradient noise in \cite{bottou2010large,ajalloeian2020convergence,faw2022power}, Assumption~\ref{assum:noise} implies that $\eta^k$  (the perturbation of a noisy gradient from a true gradient) has the magnitude depending on the size of the true gradient norm $\nabla f(x^k)$ and on the constant term $C$. 
Assumption~\ref{assum:noise} captures many noisy gradients of interest, including many biased gradients such as gradient compression, gradient smoothing, and inexact gradient oracle as shown in \cite{ajalloeian2020convergence}, and later discussed in Section~\ref{sec:App}. It also applies to machine learning problems with feature noise (e.g., missing features) \cite{khani2020feature}, robust linear regression \cite{xu2008robust}, and multi-layer neural network training with model parameters perturbed by multiplicative noise \cite{faw2022power}.
Note that Assumption~\ref{assum:noise} recovers the standard bounded variance assumption for analyzing unbiased stochastic gradient methods \cite{feyzmahdavian2016asynchronous} when we let $B=0$, and the noise assumption (4.3) in \cite{bertsekas2000gradient} when $B=C$.
We are now ready to state our main result. 
\begin{theorem}\label{thm:NCVX}
	Consider the momentum methods  in Eq. \eqref{eqn:momentum_equivalent_x_k} and \eqref{eqn:momentum_equivalent_v_k}  for   Problem \eqref{eqn:Problem}. Assume $\eta^k$ in \eqref{eqn:generic_eta_k}
 satisfies Assumption~\ref{assum:noise}.
	\begin{enumerate}
		\item \label{thm:General_NCVX} (Non-convex problems) Let Assumption~\ref{assum:smooth_bounded} hold. If $(1-\beta^2/2) B \leq 1/4$ and $0<\gamma \leq \frac{1}{L(\sqrt{\alpha} + 1)}$ with $\alpha=\frac{2(1-\beta)(\beta+2)}{\beta^2}$, then 
		\begin{align*}
			\min_{0 \leq k \leq K-1}\mathbf{E}\| \nabla f( x^k) \|^2 
			& \leq \frac{\Theta_0}{K} +4(1-\beta^2/2)C ,
		\end{align*}
  where $\Theta_0 = \frac{4}{\gamma}(f(x^0)-f^\star) + \frac{4(1-\beta)}{\beta}\| \nabla f(x^0)-v^{-1} \|^2$.
		\item \label{thm:PL_NCVX} ($\mu$-PL problems) Let Assumptions~\ref{assum:smooth_bounded} and~\ref{assum:PL} hold.  If $(2-\beta/2-\beta^2) B \leq 1/4$ and $0 < \gamma \leq \min\left( \frac{1}{L (\sqrt{ \alpha} + 1)} , \frac{\beta}{2\mu} \right)$ with $\alpha=\frac{4(1-\beta)(\beta+2)}{\beta^2}$, then 
		\begin{align*}
			\mathbf{E}[\phi^{K}] 
			& \leq (1-\mu\gamma/2)^K \phi^0 + \frac{2}{\mu}(2-\beta/2 - \beta^2) C,
		\end{align*}	
		where $\phi^k:=f(x^k)-f^\star + \frac{2\gamma(1-\beta)}{\beta} \| \nabla f(x^k)- v^{k-1} \|^2$.
	\end{enumerate}	
\end{theorem}
\begin{proof}
See Section~\ref{app:thm:General_NCVX}.
\end{proof}	

Theorem~\ref{thm:General_NCVX} establishes worst-case bounds in expectation (rather than almost-sure or in-probability) for biased momentum methods in parallel settings. 
To be more specific, for the iterates \( \{x^k\} \) with \( k < K \) and any \( K \geq 0 \), the method ensures an upper bound on the minimum gradient norm in expectation, i.e., $\min_{0 \leq k\leq K-1} \mathbf{E}\| \nabla f(x^k)\|^2$, with an $\mathcal{O}(1/K)$ rate, up to the additive constant $4(1 - \beta^2/2)C$, for general non-convex problems. 
However, this result does not guarantee that \(\mathbf{E}\| \nabla f(x^k)\|^2\) converges or that the iterate at \( k = K \) meets this bound. It only indicates that at some point within the first \( K \) iterations, the trajectory approaches a neighborhood of a stationary point without ensuring it remains there as \( k \rightarrow \infty \).
On the other hand, for problems satisfying the $\mu$-PL condition, Theorem~\ref{thm:General_NCVX}.\ref{thm:PL_NCVX} assures that the non-negative Lyapunov function in expectation, i.e., $\mathbf{E}[\phi^K]$, converges at an $\mathcal{O}((1-\mu\gamma/2)^K)$ rate to a stationary point, up to the constant error $2(2-\beta/2-\beta^2)C/\mu$.
\begin{remark} 
For $\mu$-strongly convex and smooth objective functions, which can be viewed as a special case of $\mu$-PL problems where the function is also convex, Theorem~\ref{thm:General_NCVX}.\ref{thm:PL_NCVX} ensures non-asymptotic convergence of the expected objective values to a region around the optimal value $f^{\star}$, with the size of this region growing linearly with the bias term $C$.
This result follows from two key facts: (i) the inequality $\mathbf{E}[f(x^k)-f^\star] \leq \mathbf{E}[\phi^k]$, as the second term of $\phi^k$ (i.e., $2\gamma(1-\beta) \| \nabla f(x^k)- v^{k-1} \|^2 / \beta$), is always non-negative; and (ii) the strong convexity guarantees the existence of a unique global optimal solution. 
\end{remark}
\begin{remark}
If there is no bias (i.e. $C=0$), the method requires $K \geq {\Theta_0}/{\epsilon}$ iterations to achieve $\min_{0 \leq k \leq K-1}\mathbf{E}\| \nabla f( x^k) \|^2 \leq \epsilon$ for general non-convex problems, and $K \geq 2\log\left(\phi^0/\epsilon\right)/{\mu\gamma}$ iterations to reach $\mathbf{E}[\phi^K] \leq \epsilon$ for $\mu$-PL problems.
\end{remark}

\section{Applications}\label{sec:App}

\noindent In this section, we illustrate how to use our analysis framework to establish convergence results in three main examples of biased gradients: (1) compressed gradients, (2) clipped gradients, and (3) composite gradients, including meta-learning.

\subsection{Compressed Momentum Methods}\label{subsec:comp_momen}
\noindent Communication can become the performance bottleneck of distributed optimization algorithms, especially when the network has high latency and low communication bandwidth and when the communicated information (e.g. gradients and/or model parameters) is huge, often with millions of
parameters. 
This is apparent, especially in training deep neural networks, such as AlexNet, VGG, and ResNet, where gradient communication accounts for up to $80\%$ of the running time~\cite{alistarh2017qsgd}. 
To reduce communications, several works have proposed compression operators (e.g. sparsification and/or quantization) that are applied to the information before it is communicated. 

To this end, we consider compressed momentum methods. At each iteration $k$, each worker node computes and transmits compressed stochastic gradients $Q(g_i^k)$. The server then aggregates the gradients from the worker nodes and updates the iterate $x^k$ according to:
\begin{align}
	x^{k+1} = x^k - \gamma v^k, \quad\text{where} \quad 	v^k  = v^{k-1} + \beta \left( g^k_Q- v^{k-1} \right),\label{eqn:momentumcompressed}
\end{align}
where $g^k_Q = ({1}/{n})\sum_{i=1}^n Q(g_i^k)$, $\gamma>0$, and $0\leq \beta\leq1$. Here, we assume that $g_i^k$ satisfies the variance-bounded condition
\begin{align}\label{eqn:stochasticgrad}
 \mathbf{E}\| g_i^k - \nabla f_i(x^k) \|^2 \leq \sigma^2,
\end{align}
and the compression $Q(\cdot)$ is $\alpha$-contractive with $0 < \alpha \leq 1$, i.e. 
\begin{align}\label{eqn:Compressed_def}
\| Q(x) - x \|^2 \leq (1-\alpha) \|x \|^2, \quad  \forall x\in\mathbb{R}^d.
\end{align}
Note that Eq. \eqref{eqn:stochasticgrad} and \eqref{eqn:Compressed_def} do not require the unbiasedness of stochastic gradients and compression, respectively. The $\alpha$-contractive property of $Q(\cdot)$ covers many popular deterministic compressors, e.g. Top-$K$ sparsification \cite{richtarik2021ef21,karimireddy2019error} and the scaled sign compression \cite{karimireddy2019error}.
Also notice that compressed momentum method~\eqref{eqn:momentumcompressed} is equivalent to the momentum  method in Eq.~\eqref{eqn:momentum_equivalent_x_k} and~\eqref{eqn:momentum_equivalent_v_k}  where $\eta^k=  (1/n)\sum_{i=1}^n Q(g_i^k) - \nabla f(x^k)$. We also recover compressed (stochastic) gradient descent \cite{alistarh2017qsgd,beznosikov2020biased,khirirat2021flexible}, when we let $\beta=1$.

We can now use our framework to establish the convergence of compressed momentum methods. 
We obtain the convergence from Theorem~\ref{thm:NCVX} and the next proposition. 

\begin{proposition}\label{prop:compression}
Consider the compressed momentum methods~\eqref{eqn:momentumcompressed}. Let $\mathbf{E}\| g_i^k - \nabla f_i(x^k)\|^2 \leq \sigma^2$, $\| \nabla f_i(x) - \nabla f(x)\|^2 \leq \delta^2$ for $x\in\mathbb{R}^d$,  and $\| Q(x) - x \|^2 \leq (1-\alpha) \|x \|^2$ for $0<\alpha \leq 1$ and $x\in\mathbb{R}^d$. Then, 
\begin{align*}
	\mathbf{E}\|  \eta^k \|^2 \leq B \mathbf{E}\| \nabla f(x^k) \|^2 + C,
\end{align*}	
where $B = 1-\alpha/8$, $C=(1-\alpha/4)(1+8/\alpha)\delta^2 + [(1-\alpha/2)(1+4/\alpha) + (1+2/\alpha)] \sigma^2$, and $\eta^k = (1/n)\sum_{i=1}^n Q(g_i^k) - \nabla f(x^k)$. 
\end{proposition}
\begin{proof}
See Section~\ref{app:prop:compression}.
\end{proof}

From Proposition~\ref{prop:compression} and Theorem~\ref{thm:NCVX}, we establish sublinear and linear convergence with the residual error for compressed momentum methods for general non-convex and $\mu$-PL problems, respectively.
This residual error term results from the variance $\sigma^2$ of the stochastic gradients, the compression accuracy $\alpha$, and the data similarity $\delta^2$ between the local gradient and the global gradient. 
In the centralized case (when $n=1$), the compressed momentum methods attain the $\mathcal{O}(1/{\epsilon})$ iteration complexity similarly to GD with Top-$K$ sparsification for general non-convex problems in \cite{khirirat2021flexible}, and also enjoy the $(2/(\mu\gamma))\log\left( \phi^0/\epsilon \right)$ iteration complexity for strongly convex problems analogously to GD with Top-$K$ sparsification in Section 5.1 of \cite{ajalloeian2020convergence} and with the contractive compression in Theorem 14 of \cite{beznosikov2020biased}.

\subsection{Clipped Momentum Methods}
\noindent A clipping operator is commonly used to stabilize the convergence of gradient-based algorithms for deep neural network applications. 
For example, when training recurrent neural network models in language processing, clipped gradient descent can deal with their inherent gradient explosion problems \cite{pascanu2013difficulty}, and outperforms classical gradient descent \cite{zhang2019gradient,zhang2020adaptive}.

To further improve the convergence of clipped gradient descent, we consider clipped  momentum methods that update the iterate $x^k$ via: 
\begin{align}
	x^{k+1} = x^k - \gamma v^k, ~\text{where}~ 	v^k  = v^{k-1} + \beta \left( g_c^k - v^{k-1} \right),\label{eqn:momentumclipped}
\end{align}
where $g_c^k=\frac{1}{n}\sum_{i=1}^n \mathbf{clip}_\tau(g_k^i)$,  $\gamma>0$, $0\leq \beta\leq1$, and $\mathbf{clip}_\tau(g)=\min(1,\tau/\|g\|)g$ is a clipping operator with a clipping threshold $\tau>0$. 
Unlike momentum clipping methods \cite{zhang2020improved}, our clipped momentum methods update $v^k$ based on clipped gradients instead of clipping $v^k$ and can be easily implemented in the distributed environment (where the server computes \eqref{eqn:momentumclipped} based on clipped gradients $\mathbf{clip}_\tau(g_k^i)$ aggregated from all worker nodes). 
Further note that clipped momentum methods~\eqref{eqn:momentumclipped} are momentum methods in Eq.~\eqref{eqn:momentum_equivalent_x_k} and~\eqref{eqn:momentum_equivalent_v_k} with $\eta^k=  (1/n)\sum_{i=1}^n \mathbf{clip}_\tau(g_k^i) - \nabla f(x^k)$, and also recover clipped (stochastic) gradient descent \cite{zhang2019gradient,zhang2020adaptive,mai2021stability} when we let $\beta=1$.

The convergence for Algorithm~\eqref{eqn:momentumclipped} can be obtained by Theorem~\ref{thm:NCVX} and the next proposition, which bounds  $\mathbf{E}\| \eta^k \|^2$. 
\begin{proposition}\label{prop:clip}
Consider the clipped momentum methods~\eqref{eqn:momentumclipped}. Let  $f_i(x)$ have $L$-Lipschitz continuous gradient, $\mathbf{E}\| g_i^k - \nabla f_i(x^k)\|^2 \leq \sigma^2$, and $f(x)-f(x^\star) \leq \delta$ for $\delta>0$ and $x\in\mathbb{R}^d$. Then, 
\begin{align*}
	\mathbf{E}\|  \eta^k \|^2 \leq C,
\end{align*}	
where 
$C= \max(2\sigma^2 + 4L\delta + \tau^2, 0 ) + 2\sigma^2$,  and $\eta^k = \frac{1}{n}\sum_{i=1}^n \mathbf{clip}_\tau(g_k^i) - \nabla f(x^k)$. 
\end{proposition}
\begin{proof}
See Section~\ref{app:prop:clip}.
\end{proof}

Similarly to Section~\ref{subsec:comp_momen}, we obtain the unified convergence for clipped momentum methods by using Proposition~\ref{prop:clip} and Theorem~\ref{thm:NCVX}. 
The residual error term from clipped momentum methods comes from the bias introduced by gradient clipping $\tau$.
If we choose $v^{-1}=\nabla f(x^0)$ and $\tau = \sqrt{(1+\theta)M^2 + (1+1/\theta)\sigma^2}$ for $\theta>0$, then we have $\mathbf{E}\| \eta^k \|^2 = 0$. 
Clipped momentum methods thus reach the iteration complexities of ${\Theta_0}/{\epsilon}$ to achieve $\min_{0 \leq k \leq K-1}\mathbf{E}\| \nabla f( x^k) \|^2 \leq \epsilon$ for general non-convex problems, and
$2\log\left(\phi^0/\epsilon\right)/{\mu\gamma}$ iterations to reach $\mathbf{E}[\phi^K] \leq \epsilon$ for $\mu$-PL problems.
In particular, for general non-convex problems, clipped momentum methods attain a lower iteration complexity than momentum clipping methods \cite{zhang2020improved} (with $\mathcal{O}(\epsilon^{-4})$) at the price of a more restrictive condition on the gradient-bounded norm.   

\subsection{Stochastic Composite Momentum  Methods}

\noindent Finally, we turn our attention to the problem of minimizing functions with composite finite sum structure, i.e. Problem \eqref{eqn:Problem} with  $f_i(x)= F_i\left( g_i(x) \right)$, $F_i(x) = \frac{1}{m_F}\sum_{j=1}^{m_F} F_{i,j}(x)$, and $g_i(x) = \frac{1}{m_g}\sum_{j=1}^{m_g} g_{i,j}(x)$.
Such composite finite-sum minimization problems arise in many applications such as reinforcement learning \cite{huo2018accelerated}, multi-stage stochastic programming \cite{shapiro2021lectures}, and risk-averse portfolio optimization \cite{ravikumar2009sparse}.

Another instance of a problem with this structure is Model-Agnostic Meta-Learning (MAML) \cite{finn2017model,fallah2020personalized}. Unlike traditional learning, MAML focuses on finding a model that performs well across multiple tasks. MAML considers a scenario where the model is updated based on small data partitions from every new task, typically by using just one or a few gradient descent steps on each task.
In distributed MAML, assuming each worker node takes the initial model and performs a single gradient descent step using its private loss function, we can reformulate the problem described in Eq.~\eqref{eqn:Problem} as a distributed MAML problem
\begin{align}\label{eqn:prolem_MAML}
     \mathop{\text{minimize}}\limits_{x\in\mathbb{R}^d} \quad \frac{1}{n}\sum_{i=1}^n f_i(x - \gamma \nabla f_i(x)),
\end{align}
where $\gamma>0$ is the positive step-size, $f_i(x)=\frac{1}{m}\sum_{j=1}^{m} \ell_{i}(x,a^j_i,b^j_i)$ is the loss function privately known by worker node $i$, and $a_i^j\in\mathbb{R}^d$ is the $j^{\rm th}$ data sample belonging to worker node $i$ with its associated class label $b_i^j \in \{-1,1\}$ for $j=1,2,\ldots,m$.
Here, we assume that the entire dataset is distributed evenly among $n$ worker nodes. 
Also note that this MAML problem \eqref{eqn:prolem_MAML} is thus the composite finite-sum minimization problem over $\frac{1}{n}\sum_{i=1}^n F_i(g_i(x))$ with $F_i(x) =f_i(x)$, $F_{i,j}(x)=\ell_{i}(x,a^j_i,b^j_i)$, $g_i(x)=x-\gamma\nabla f_i(x)$, $g_{i,j}(x)=x-\gamma \nabla \ell_{i}(x,a^j_i,b^j_i)$, and $m_F=m_g = m$.

To solve the composite finite-sum minimization problem, we consider  stochastic composite momentum methods which update  $x^k$ via: 
\begin{align}
	x^{k+1} = x^k - \gamma v^k, \quad\text{where} \quad 	v^k  = v^{k-1} + \beta \left( \tilde \nabla f(x^k) - v^{k-1} \right),\label{eqn:momentumComposite}
\end{align}
where 
\begin{align}\label{eqn:Composite_stochGrad}
\tilde \nabla f(x^k) & = \frac{1}{n}\sum_{i=1}^n \tilde \nabla f_i(x^k) \\ 
  \tilde{\nabla} f_i(x^k) &= \left\langle  \nabla g_{\mathcal{S}_g}(x^k),  \frac{1}{\vert \mathcal{S}_F^k\vert} \sum_{j\in\mathcal{S}_F^k} \nabla F_{i,j} \left( g_{\mathcal{S}_F}(x^k) \right) \right\rangle  
\end{align}
and also $\nabla g_{\mathcal{S}_g}(x^k)=\frac{1}{\vert \mathcal{S}_g^k \vert} \sum_{j\in\mathcal{S}^k_g} \nabla g_{i,j}(x^k)$ and $g_{\mathcal{S}_F}(x^k)=\frac{1}{\vert \mathcal{S}^k_g \vert} \sum_{j\in\mathcal{S}_g^k} g_{i,j}(x^k)$.
Here, $\mathcal{S}_g^k$ and $\mathcal{S}_F^k$ are the subsets sampled from the set 
$\{g_{i,1},\ldots, g_{i,m_g}\}$ and $\{ F_{i,1},\ldots, F_{i,m_F}\}$, respectively, uniformly at random. 
Note that $\tilde{\nabla} f_i(x^k)$ is the biased estimator of the full  gradient: 
\begin{align}\label{eqn:Composite_fullGrad}
    \nabla f_i(x^k) = \left\langle \nabla g_i(x^k),  \nabla F_i \left( g_i(x^k) \right) \right\rangle.
\end{align}
We can thus show that Algorithm~\eqref{eqn:momentumComposite} is the momentum method in Eq.~\eqref{eqn:momentum_equivalent_x_k} and~\eqref{eqn:momentum_equivalent_v_k}  with $\eta^k = (1/n)\sum_{i=1}^n [\nabla \tilde f_i(x^k) - \nabla f_i(x^k)]$ and  prove the upper bound for $\mathbf{E}\|\eta^k \|^2$ as following. 

\begin{proposition}\label{prop:stochastic_composite}
Consider the stochastic composite momentum methods~\eqref{eqn:momentumComposite} for the problem of minimizing $f(x)= \frac{1}{n}\sum_{i=1}^n F_i \left( g_i(x) \right)$, where $F_i(x) = \frac{1}{m_F}\sum_{j=1}^{m_F} F_{i,j}(x)$ and $g_i(x) = \frac{1}{m_g}\sum_{j=1}^{m_g} g_{i,j}(x)$. Let each $F_{i,j}(x)$ be $\ell_F$-Lipschitz continuous and have $L_F$-Lipschitz continuous gradient, and each $g_{i,j}(x)$ be $\ell_g$-Lipschitz continuous and have $L_g$-Lipschitz continuous gradient. 
Also suppose that $\mathbf{E}\| g_{i,j}(x) - g_i(x) \|^2 \leq \sigma_g^2$,  $\mathbf{E}\| \nabla g_{i,j}(x) - \nabla g_i(x) \|^2 \leq \sigma_{\nabla g}^2$, $\mathbf{E}\| \nabla F_{i,j}(x) - \nabla F_i(x) \|^2 \leq \sigma_F^2$, $\vert \mathcal{S}_g^k \vert=S_g$, and $\vert \mathcal{S}_F^k \vert=S_F$. 
Further denote $\eta^k = \frac{1}{n}\sum_{i=1}^n [\nabla \tilde f_i(x^k) - \nabla f_i(x^k)]$, where $\tilde \nabla f_i(x^k)$ and $\nabla f_i(x^k)$ are defined in \eqref{eqn:Composite_stochGrad} and \eqref{eqn:Composite_fullGrad}.
Then, 
\begin{enumerate}
    \item $f(x)$ has $L$-Lipschitz continuous gradient with $L=L_g \ell_F + \ell_g^2 L_F$.\label{prop:stochastic_composite_1}
    \item $\mathbf{E}\| \eta^k \|^2 \leq  C$ with  $C= 3 \ell_g^2 \frac{\sigma_F^2}{S_F}   + 3 \ell_F^2 \frac{\sigma_{\nabla g}^2}{S_g}   + 3 \ell_g^2 L_F^2 \frac{\sigma_g^2}{S_g}$. \label{prop:stochastic_composite_2}
\end{enumerate}
\end{proposition}
\begin{proof}
See Section~\ref{app:prop:stochastic_composite}.
\end{proof}

From Proposition~\ref{prop:stochastic_composite} and Theorem~\ref{thm:NCVX} we establish the convergence for stochastic composite momentum methods for general non-convex problems. 
SCGD from Theorem 8 of \cite{wang2017stochastic}.
To reach $\min_{0 \leq k \leq K-1}\mathbf{E}\| \nabla f( x^k) \|^2 \leq \epsilon$, stochastic composite momentum methods with $S_F=S_g=S$ need $\mathcal{O}(\epsilon^{-1})$ iterations when $S=8(1-\beta^2/2)(3\ell_g^2 \sigma_F^2 + 3\ell_F^2 \sigma_{\nabla g}^2   + 3 \ell_g^2 L_F^2  \sigma_{g}^2)/{\epsilon}$. 
This iteration complexity is lower than the $\mathcal{O}(\epsilon^{-4})$ iteration complexity by SCGD from Theorem 8 of \cite{wang2017stochastic}.
These results drawn from Proposition~\ref{prop:stochastic_composite} and Theorem~\ref{thm:NCVX} can be indeed applied to the MAML problem \eqref{eqn:prolem_MAML} as shown in the proposition below:
\begin{proposition}\label{prop:MAML_SC}
Consider the MAML problem \eqref{eqn:prolem_MAML}. Let $\ell_{i}(x,a^j_i,b^j_i)$ be $\ell_l$-Lipschitz continuous and have $L_l$-Lipschitz continuous gradient. Then, the MAML problem is  the problem of minimizing $f(x)= \frac{1}{n}\sum_{i=1}^n F_i \left( g_i(x) \right)$, where $F_i(x) = \frac{1}{m}\sum_{j=1}^{m} F_{i,j}(x)$, $g_i(x) = \frac{1}{m}\sum_{j=1}^{m} g_{i,j}(x)$, $F_{i,j}(x)=\ell_{i}(x,a^j_i,b^j_i)$, and $g_{i,j}(x)=x-\gamma \nabla \ell_{i}(x,a^j_i,b^j_i)$. In addition, each $F_{i,j}(x)$ is $\ell_F$-Lipschitz continuous with $\ell_F=\ell_l$ and has $L_F$-Lipschitz continuous gradient with $L_F=L_l$, and also each $g_{i,j}(x)$ is $\ell_g$-Lipschitz continuous with $\ell_g=1+\gamma L_l$ and has $L_g$-Lipschitz continuous gradient with $L_g=2 \gamma L_l$.
\end{proposition}
\begin{proof}
See Section~\ref{app:prop:MAML_SC}.    
\end{proof}

\section{Numerical Experiments}\label{sec:exp}
\subsection{Distributed Deep Neural Networks}
\noindent To demonstrate the superior performance of momentum methods over SGD, we evaluated both methods using biased gradient estimators on training deep neural network models over the MNIST and FashionMNIST datasets.

\subsubsection{Fully connected neural networks over MNIST}
\noindent The MNIST dataset consists of 60000 training images and 10000 test images. Each image is in a 28 $\times$ 28 gray-scale and represents one of the digits from $0$ to $9$ \cite{lecun1998mnist}. For this task, we use a fully connected neural network (FCNN) with two hidden layers. Each hidden layer has $512$ neurons and the ReLU activation function. This FCNN architecture thus has 669706 trainable parameters, and is trained to solve the problem of minimizing the cross-entropy loss function. 

\subsubsection{ResNet-18 networks over FashionMNIST}
\noindent We also examine the FashionMNIST dataset, which has the same number of samples and of classes as MNIST.
Since training over FashionMNIST is more challenging than MNIST \cite{xiao2017fashion}, it is recommended to use convolution neural networks. In particular, we employ the well-known 18-layer Residual Network model (i.e. ResNet-18) \cite{he2016deep}. By adapting the number of model outputs to 10 classes, the number of trainable parameters of this ResNet-18 model is 11181642.

The numerical experiments for both cases above are implemented in Python 3.8.6 on a computing server with NVIDIA Tesla T4 GPU with 16GB RAM. The weights of the neural networks are initialized by the default random initialization routines of the Pytorch framework.
For SGD, we evaluate various fixed step-sizes; $\gamma \in \{0.01, 0.05, 0.1, 0.3, 0.5, 0.7\}$. In the case of momentum methods (which we call SGDM to distinguish it from SGD), we consider $\beta \in \{0.01, 0.1, 0.3, 0.6, 0.9\}$ under the same choice of step-size as SGD. We do the experiments in a server-worker setup with $20$ and $100$ workers. 

To study the performance of SGD and SGDM under biased gradient estimation, we investigate the clipping operator and Top-$K$ sparsification. For the clipping operator, we evaluate three different $\tau \in \{1, 2, 5\}$ for both problems. However, in the case of  Top-$K$ sparsification, since the number of trainable parameters of FCNN and ResNet-18 is not in the same order, different $K$ have been used. Specifically, we choose  $K \in \{13400, 33500, 67000\}$ for the FCNN model over MNIST dataset, and $K \in \{11181, 111816, 1118164\}$ for the ResNet-18 network over FashionMNIST dataset. 
For brevity, we present the ratio of $K$ to the trainable parameters (in percent) rather than the value of $K$ for the rest of this section.

Due to the stochastic nature of neural networks, we repeated our experiments with five different random seeds for network initialization to ensure a fair comparison between the SGDM and SGD. 
We monitor both training loss and test accuracy and plot the average of these metrics alongside their standard deviation across different trials. 
Learning curves of SGDM and SGD on the MNIST and FashionMNIST datasets are shown in Figure \ref{fig1}. A more thorough evaluation of the considered hyperparameters on both tasks can be found in Appendix \ref{AdditionalCurves}.

\subsubsection{Result discussions.}
\begin{figure*}[ht] 
     \centering
     \subfloat[Top-2\% sparsification]{
         \includegraphics[width=0.41\linewidth] {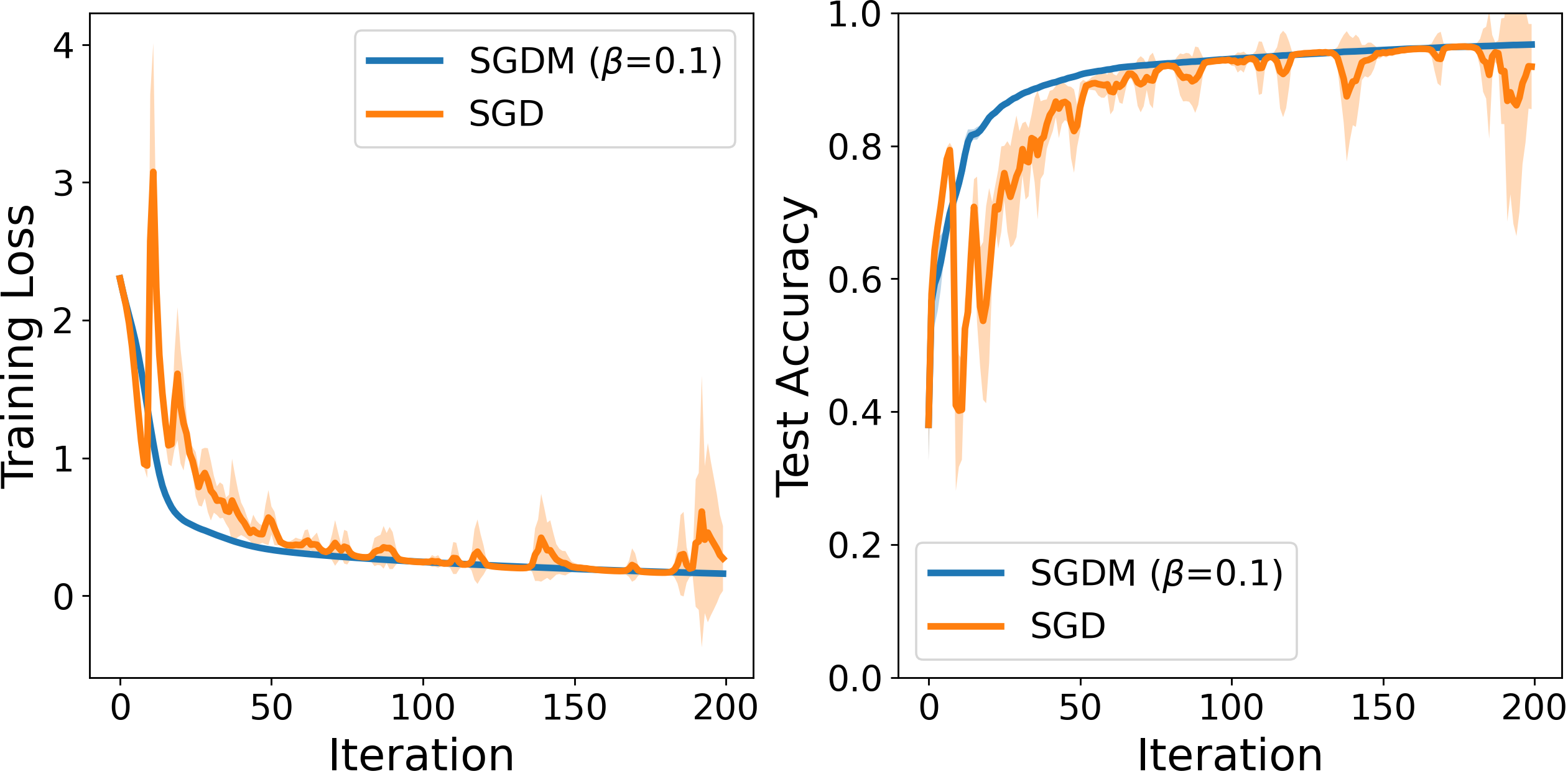}
         \label{fig1a}} 
     \subfloat[Clipped with $\tau=2$]{
         \includegraphics[width=0.41\linewidth]{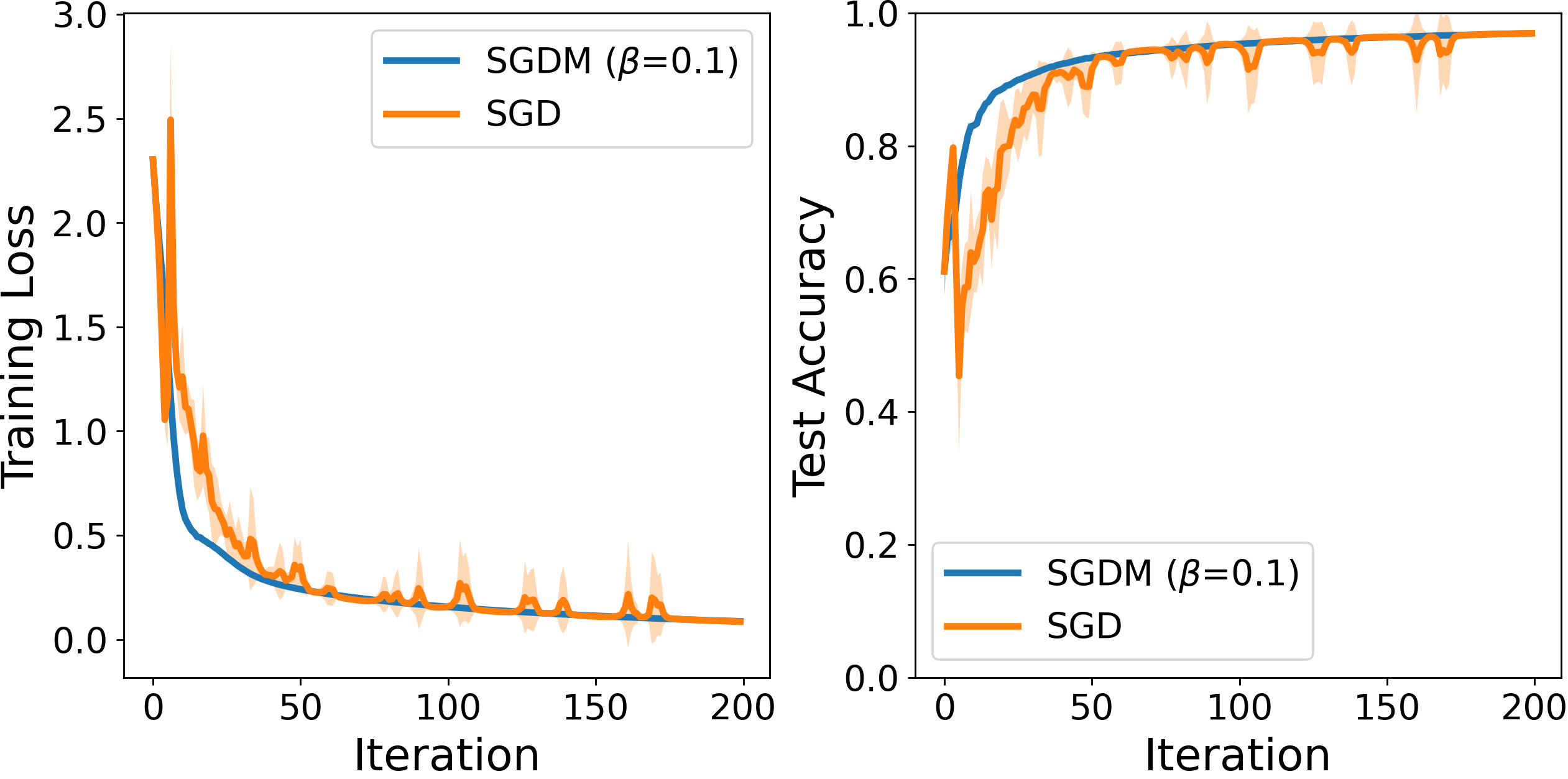}
         \label{fig1b}}
     \hfill   
     \subfloat[Top-0.1\% sparsification]{
        \includegraphics[width=0.41\linewidth]{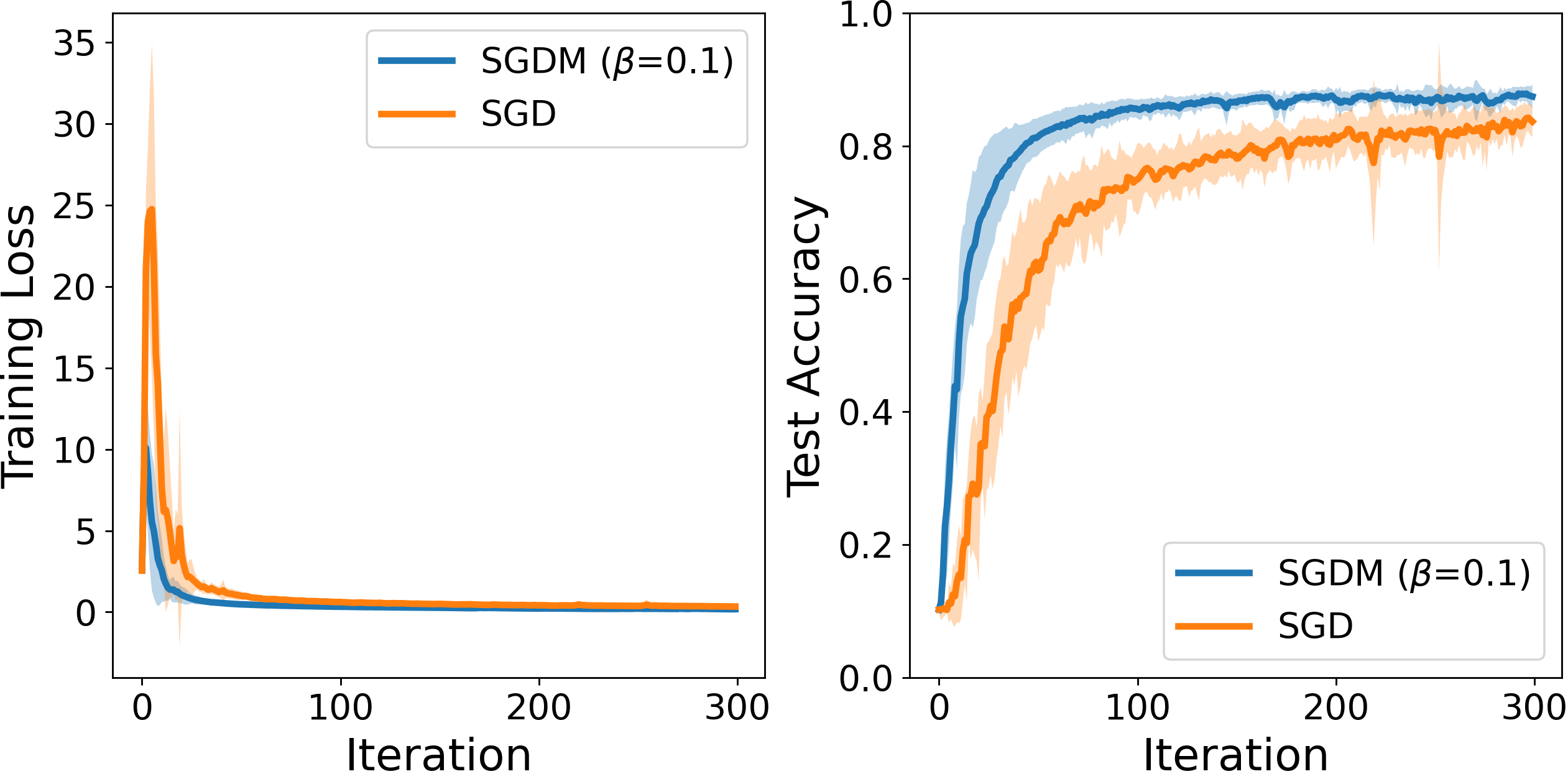}
         \label{fig2a}}
     \subfloat[Clipped with $\tau=2$]{
         \includegraphics[width=0.41\linewidth]{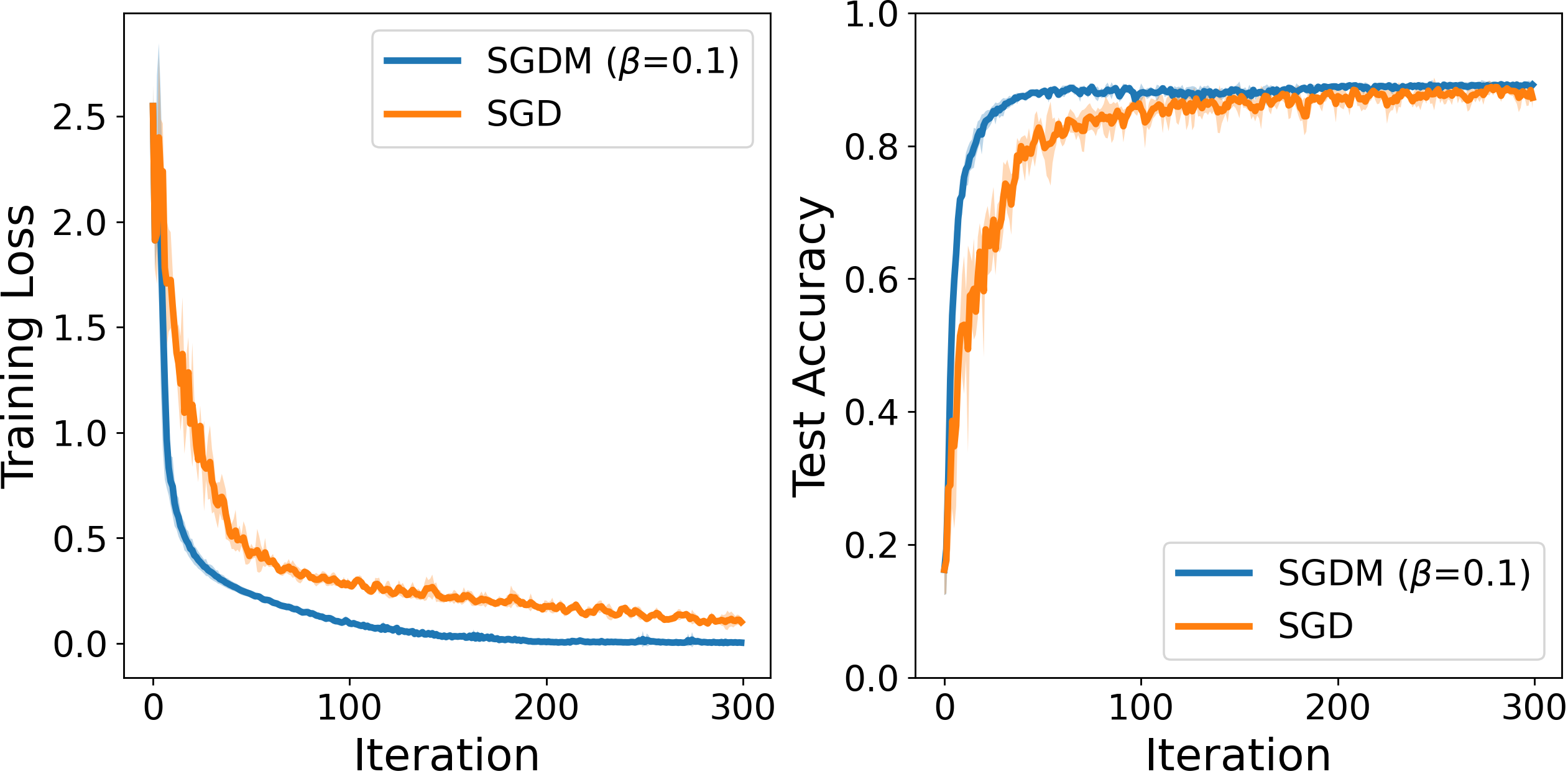}
         \label{fig2b}}
        \caption{Performance of  parallel SGDM (i.e.  momentum method) and SGD method under various biased gradient estimations ((a) and (c) compressed gradients; (b) and (d) clipped gradients) in terms of (left plots -) training loss and (right plots -) test accuracy on (top plots -) MNIST and (bottom plots -) FashionMNIST datasets, considering $n=100$ and $\gamma=0.5$. }
        \label{fig1}
        \label{fig2}
\end{figure*}

\noindent Regardless of the model and dataset, under any combination of the mentioned hyperparameters, we consistently see a better performance of SGDM than SGD under biased gradients (either the Top-$K$ sparsification or the clipping operator) in terms of both training loss and test accuracy.
Particularly, we make the following observations:
First, SGDM converges faster than SGD, while sometimes SGD even diverges for large step-sizes 
(See Appendix \ref{AdditionalCurves}).
Second, SGDM is more stable and has less extreme fluctuations.
Third, SGDM is more robust to different network initialization and has higher performance reliability in different trials, as it consistently achieves high convergence speed and low variance.
Fourth, SGDM is less sensitive to the degree of gradient bias (precisely by the value of $K$ and $\tau$) 
(See, e.g. Figure \ref{fig1}).
Fifth, in many cases, SGDM leads to convergence to less loss and higher accuracy. For example, in Figure \ref{fig2a}, the loss and accuracy of SGDM are 0.17 and 0.87, respectively, while for SGD, they are 0.34 and 0.83 at iteration 300.

\subsection{Experimental Validation of Theoretical Findings}

 \begin{figure*}[ht] 
     \centering
     \subfloat[Effect of the parameter $K$ ]{
         \includegraphics[width=0.315\linewidth]{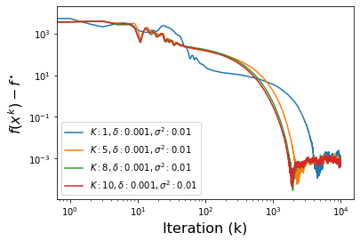}
         \label{figT.a}}
     \subfloat[Effect of the parameter $\delta$]{
         \includegraphics[width=0.315\linewidth]{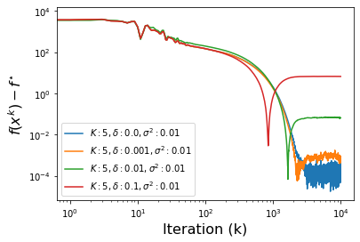}
         \label{figT.b}}
     \hfill   
     \subfloat[Effect of the parameter $\sigma^2$]{
        \includegraphics[width=0.315\linewidth]{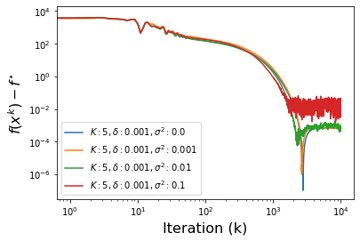}
         \label{figT.c}}
        \caption{Effect of the parameters $\sigma^2$, $\delta$, and $K$, which change the noise, bias, and compression level, respectively, for Top-$K$ sparsification. Here we optimize $f(x)= \frac{1}{2}||Ax||^2, ~x \in \mathbb{R}^{10}$, considering $\gamma = 0.5$, and $\beta = 0.1$.}
        \label{figT}
\end{figure*}
\noindent We conduct a numerical investigation to ascertain the alignment between our theoretical bounds and the practical performance of SGDM 
under biased gradient estimation. Our analysis mirrors the approach outlined in \cite{ajalloeian2020convergence}.
Our experimentation encompasses scenarios where we introduce a constant offset \( \delta \) to the gradient to regulate bias, alongside varying levels of noise \( \sigma^2 \). Additionally, we explore implicit biases stemming from gradient compression via the Top-\(K\) sparsification technique, i.e.  Top-\(K\big(\nabla f(x) + \delta + \mathcal{N}(0, \sigma^2)\big)\).
\subsubsection{Result Discussions} Figures \ref{figT.a}, \ref{figT.b}, and \ref{figT.c} depict the influence of parameters \( K \), \( \delta \), and \( \sigma^2 \) on the convergence rate and error floor when employing Top-\(K\) compressors alongside constant \( \gamma \) and \( \beta \).
It's noteworthy that a minor bias only impacts the convergence neighborhood when stochastic noise is absent, corroborating Proposition~\ref{prop:compression}. When \( \delta \) is insignificantly small compared to \( \sigma^2 \), i.e.  when bias pales against noise, the second term of \( C \) in Proposition~\ref{prop:compression} dictates the proximity of the optimal solution to which SGDM converges with a constant step-size. Conversely, a substantial \( \delta \) tilts the balance towards the dominance of the first term in \( C \), thus determining the convergence neighborhood. As elucidated in Theorem~\ref{thm:General_NCVX}, the second term's dependency on \( \beta \) implies that adjusting momentum weight can regulate residual error, a capability not afforded by SGD.
From Figure \ref{figT.a}, it's apparent that under various compression levels, Top-\(K\) achieves convergence to nearly the same level as without compression, albeit at a slower pace. This resonates with Proposition~\ref{prop:compression}, where we observe \( \alpha \) influencing both terms of \( C \).
The convergence rate of compression methods decelerates with decreasing compression parameter \( K \), mirroring \( B \) in Proposition~\ref{prop:compression}.

\section{Proofs}

\subsection{Proof of Lemma \ref{lemma:descent}}\label{app:lemma:descent}
\noindent We prove the result in two steps. 

\paragraph*{\bf Step 1) The upper bounds for $\| \nabla f(x^k) - v^k \|^2$ and $\| \nabla f(x^{k+1}) - v^k \|^2$} First, we bound $\| \nabla f(x^k) - v^k \|^2$. By the convexity of the squared norm, for $0\leq \beta \leq 1$,  
\begin{align}\label{eqn:descent_interm_x_v}
\| \nabla f(x^k) - v^k \|^2 \leq (1-\beta)\| \nabla f(x^k) - v^{k-1} \|^2 + \beta \|  \eta^k \|^2,
\end{align}
where $\eta^k = (1/n)\sum_{i=1}^n  \tilde \nabla f_i(x^k) - \nabla f(x^k)$. Second, we bound $\| \nabla f(x^{k+1}) - v^k \|^2$ as follows:
\begin{align*}
	& \| \nabla f(x^{k+1}) - v^k \|^2  
\mathop{\leq}^{\eqref{eqn:theta_x_y}+\eqref{eqn:descent_interm_x_v}} (1+\theta)(1-\beta)\| \nabla f(x^k) - v^{k-1} \|^2 \\
&\hspace{0.5cm}+ (1+\theta)\beta \|  \eta^k \|^2  + (1+1/\theta) \| \nabla f(x^{k+1}) - \nabla f(x^k) \|^2.
\end{align*}	

If $\theta=\beta/2$, then by the fact that $(1+\beta/2)(1-\beta) \leq 1-\beta/2$ and by  Assumption~\ref{assum:smooth_bounded}
\begin{align}
    & \| \nabla f(x^{k+1}) - v^k \|^2  
     \leq (1-\beta/2) \| \nabla f(x^k) - v^{k-1} \|^2 \notag \\
    & \hspace{0.5cm} + (1+\beta/2)\beta \| \eta^k \|^2 + (1+2/\beta)L^2 \| x^{k+1} - x^k \|^2.\label{eqn:descent_interm_x_v_2}
\end{align}	

\paragraph*{\bf Step 2)  The  inequality}
Define $\phi^k:=f(x^k)-f^\star + A \| \nabla f(x^k)- v^{k-1} \|^2$ with $A>0$. If $\gamma \leq 1/L$, then
\begin{align}
	\phi^{k+1} 
	& \mathop{\leq}^{\eqref{eqn:smoothness_ineq} + \eqref{eqn:product_eq}}  f(x^k)-f^\star - \frac{\gamma}{2}\| \nabla f(x^k) \|^2 - D_1\| x^{k+1} - x^k\|^2 \notag \\
	&\hspace{0.5cm} + \frac{\gamma}{2}\| \nabla f(x^k)- v^k \|^2 + A\|  \nabla f(x^{k+1})- v^{k} \|^2. \label{eqn:Lemma_step2_descent}
\end{align}	
where $D_1 =  \frac{1}{2\gamma} - \frac{L}{2}$. Next, 
plugging \eqref{eqn:descent_interm_x_v} and \eqref{eqn:descent_interm_x_v_2} into \eqref{eqn:Lemma_step2_descent} yields
\begin{align*}
	\phi^{k+1} 
	&\leq f(x^k)-f^\star - \frac{\gamma}{2}\| \nabla f(x^k) \|^2 + B_1 \| \nabla f(x^k)-v^{k-1} \|^2 \\
	&\hspace{0.5cm}- B_2 \| x^{k+1} - x^k \|^2 + B_3\| \eta^k\|^2,
\end{align*}	
where  $B_1 = \gamma (1-\beta)/2 + A(1-\beta/2)$, $B_2 =  \frac{1}{2\gamma} - \frac{L}{2}  - A \frac{(\beta+2)L^2}{\beta}$, and $B_3 =\gamma\beta/2 + A\beta(1+\beta/2)$.

\subsection{Proof of Theorem~\ref{thm:NCVX}}\label{app:thm:General_NCVX}
\noindent We prove the convergence in $\min_k\mathbf{E}\| \nabla f(x^k) \|^2$ for general non-convex problems in Theorem~\ref{thm:NCVX}-\ref{thm:General_NCVX}, and the convergence in $\mathbf{E}[\phi^k]$ for $\mu$-PL problems in Theorem~\ref{thm:NCVX}-\ref{thm:PL_NCVX}.

\subsubsection{Proof of Theorem~\ref{thm:NCVX}-\ref{thm:General_NCVX}}
We prove Theorem~\ref{thm:NCVX}-\ref{thm:General_NCVX} by deriving the descent inequality from \eqref{eqn:descent}. If $A = \frac{\gamma}{\beta}(1-\beta)$,
\begin{align*}
\phi^{k+1} 
&\leq \phi^k - \frac{\gamma}{2}\| \nabla f(x^k) \|^2 - \Theta\| x^{k+1} - x^k \|^2  \nonumber \\
&\hspace{0.5cm}+ \gamma (1-\beta^2/2) \| \eta^k\|^2 \\
&\leq \phi^k - \frac{\gamma}{2}\| \nabla f(x^k) \|^2 + \gamma (1-\beta^2/2) \| \eta^k\|^2,
\end{align*}	
where $\Theta = \frac{1}{2\gamma} - \frac{L}{2} - {\gamma} \cdot  \frac{(1-\beta)(\beta+2)L^2}{\beta^2} $.
Here, the second inequality comes from the fact that $\Theta \geq 0$ when $0<\gamma \leq \frac{1}{L}\cdot\frac{1}{ \sqrt{ \frac{2(1-\beta)(\beta+2)}{\beta^2}} + 1}$ by using Lemma~\ref{lemma:stepsize_range} with $a=1/2$, $b=L/2$ and $c =\frac{(1-\beta)(\beta+2)L^2}{\beta^2}$.
	
Next, by taking the expectation, using the fact that $\mathbf{E}\| \eta^k \|^2 \leq B \mathbf{E}\| \nabla f(x^k)\|^2 + C$ for $B, C \geq 0$, and assuming that $(1-\beta^2/2)B \leq 1/4$, we have 
\begin{align}
	\mathbf{E}[\phi^{k+1}] 
	&\leq \mathbf{E}[\phi^k] - \gamma D \mathbf{E}\| \nabla f(x^k) \|^2 + \gamma(1-\beta^2/2)C \nonumber \\
 &\leq \mathbf{E}[\phi^k] - \frac{\gamma}{4}  \mathbf{E}\| \nabla f(x^k) \|^2 + \gamma(1-\beta^2/2)C.\label{eqn:Theorem1_generalncvx_descent}
\end{align}	
where $D = 1/2 - (1-\beta^2/2)B$. 
 	
Next, we derive the convergence in $\min_{0 \leq k \leq K-1}\mathbf{E}\| \nabla f( x^k) \|^2$ from \eqref{eqn:Theorem1_generalncvx_descent}. 
By the fact that $\min_{0 \leq k \leq K-1}\mathbf{E}\| \nabla f( x^k) \|^2 \leq  \frac{1}{K}\sum_{k=0}^{K-1} \mathbf{E} \| \nabla f(x^k) \|^2$, by \eqref{eqn:Theorem1_generalncvx_descent}, by the consequence of telescopic series, and finally by the definition of $\phi^0$, we complete the proof.

\subsubsection{Proof of Theorem~\ref{thm:NCVX}-\ref{thm:PL_NCVX}}
We show Theorem~\ref{thm:NCVX}-\ref{thm:PL_NCVX} by proving the descent inequality in $\mathbf{E}[\phi^k]$ from \eqref{eqn:descent}. 
If $A = \frac{2\gamma}{\beta}(1-\beta)$, then
\begin{align*}
	\phi^{k+1} 
\leq&  f(x^k)-f^\star - \frac{\gamma}{2}\| \nabla f(x^k) \|^2 - \hat\Theta\| x^{k+1} - x^k \|^2  \\
& + A (1-\beta/4) \| \nabla f(x^k)-v^{k-1} \|^2 + \gamma \hat D \| \eta^k\|^2 \\
& \leq  f(x^k)-f^\star - \frac{\gamma}{2}\| \nabla f(x^k) \|^2   \\
     & + A (1-\beta/4) \| \nabla f(x^k)-v^{k-1} \|^2  + \gamma \hat D \| \eta^k\|^2,
\end{align*}
where $\hat\Theta= \frac{1}{2\gamma} - \frac{L}{2} - \gamma\frac{2(1-\beta)(\beta+2)L^2}{\beta^2}$ and $\hat D = 2-\beta/2 - \beta^2$.
The second inequality derives from the fact that $\hat\Theta \geq 0$ when $0<\gamma \leq \frac{1}{ \sqrt{ \frac{4(1-\beta)(\beta+2)L^2}{\beta^2}} + L}$ by using Lemma~\ref{lemma:stepsize_range} with $a=1/2$, $b=L/2$ and $c=\frac{2(1-\beta)(\beta+2)L^2}{\beta^2}$.

Next, by taking the expectation, using the fact that $\mathbf{E}\| \eta^k \|^2 \leq B \mathbf{E}\| \nabla f(x^k)\|^2 + C$ for  $B, C \geq 0$, and assuming that $(2-\beta/2 - \beta^2) B \leq 1/4$,
\begin{align*}
	\mathbf{E}[\phi^{k+1}] 
	  &  \leq \mathbf{E}[f(x^k)-f^\star] - \gamma\Theta_1 \mathbf{E}\| \nabla f(x^k) \|^2  \\
	&+ A (1-\beta/4) \mathbf{E} \| \nabla f(x^k)-v^{k-1} \|^2  + \gamma \hat D C \\
 & \overset{\text{Assumption~\ref{assum:PL} }}{\leq}  (1-\mu\gamma/2)\mathbf{E}[f(x^k)-f^\star]  \\
 & + A (1-\beta/4) \mathbf{E} \| \nabla f(x^k)-v^{k-1} \|^2  + \gamma \hat D C.
\end{align*}
where $\Theta_1 = 1/2 - (2-\beta/2 - \beta^2) B$.

If $\gamma \leq \frac{\beta}{2\mu}$, then we obtain the descent inequality
\begin{align*}
	\mathbf{E}[\phi^{k+1}] 
	&\leq  (1-\mu\gamma/2)\mathbf{E}[\phi^k]   + \gamma \hat D C.
\end{align*}
Finally, applying this inequality over $k=0,1,\ldots,K-1$ yields the  result.

\subsection{Proof of Proposition~\ref{prop:compression}}\label{app:prop:compression}
\noindent The compressed momentum methods~\eqref{eqn:momentumcompressed} are momentum methods in Eq.~\eqref{eqn:momentum_equivalent_x_k} and~\eqref{eqn:momentum_equivalent_v_k} where $\eta^k=  \frac{1}{n}\sum_{i=1}^n Q(g_i^k) - \nabla f(x^k)$. 
We now bound $\| \eta^k \|^2$ by using Lemma~\ref{lemma:eta_k_supp} and  \eqref{eqn:Compressed_def}.

From Lemma~\ref{lemma:eta_k_supp} with $G(g_i^k) = Q(g_i^k)$, and by  \eqref{eqn:Compressed_def}
\begin{align*}
	\| \eta^{k} \|^2 
 \leq  \frac{B_1}{n}\sum_{i=1}^n \| g_i^k\|^2   + \frac{B_2}{n}\sum_{i=1}^n \| g_i^k - \nabla f_i(x^k) \|^2,
\end{align*}	
where $B_1 = (1+\theta_1)(1-\alpha)$ and $B_2 = (1+1/\theta_1)$.

If $\theta_1=\alpha/2$, then $(1+\theta_1)(1-\alpha) \leq 1-\alpha/2$ and by \eqref{eqn:theta_x_y} with $x=\nabla f_i(x^k)$ and $y = g_i^k - \nabla f_i(x^k)$, 
\begin{align*}
	\| \eta^{k} \|^2 
	 \leq  \frac{\hat B_1}{n}\sum_{i=1}^n \| \nabla f_i(x^k)\|^2 + \frac{\hat B_2}{n}\sum_{i=1}^n \| g_i^k - \nabla f_i(x^k) \|^2,
\end{align*}	
where  $\hat B_1 = (1+\theta_2)(1-\alpha/2)$ and $\hat B_2 = (1+1/\theta_2)(1-\alpha/2)+ (1+2/\alpha)$.

If $\theta_2=\alpha/4$, then $(1+\theta_2)(1-\alpha/2) \leq 1-\alpha/4$ and by  \eqref{eqn:theta_x_y} with $x=\nabla f(x^k)$ and $y=\nabla f_i(x^k) - \nabla f(x^k)$
\begin{align*}
	\| \eta^{k} \|^2 
	\leq &   \bar B_1 \| \nabla f(x^k) \|^2 + \frac{\bar B_2}{n}\sum_{i=1}^n \| \nabla f_i(x^k) -\nabla f(x^k)\|^2  \\
	&+ \frac{\bar B_3}{n}\sum_{i=1}^n \| g_i^k - \nabla f_i(x^k) \|^2,
\end{align*}	
where $\bar B_1 = (1+\theta_3)(1-\alpha/4)$, 
$\bar B_2 = (1+1/\theta_3)(1-\alpha/4)$ and 
$\bar B_3 = (1+4/\alpha)(1-\alpha/2)+ (1+2/\alpha)$.
Finally, If $\theta_3=\alpha/8$, then $(1+\theta_3)(1-\alpha/4) \leq 1-\alpha/8$, and then by taking the expectation and by using the fact that $\mathbf{E}\| g_i^k - \nabla f_i(x^k)\|^2 \leq \sigma^2$ and that $\| \nabla f_i(x) - \nabla f(x)\|^2 \leq \delta^2$ for $x\in\mathbb{R}^d$, we complete the proof. 

\subsection{Proof of Proposition~\ref{prop:clip}}\label{app:prop:clip}
\noindent 
The compressed momentum methods~\eqref{eqn:momentumcompressed} are momentum methods in Eq.~\eqref{eqn:momentum_equivalent_x_k} and \eqref{eqn:momentum_equivalent_v_k} where $\eta^k=  \frac{1}{n}\sum_{i=1}^n \mathbf{clip}_\tau(g_k^i) - \nabla f(x^k)$. 
To bound $\| \eta^k \|^2$, 
we first introduce one useful lemma for proving the result. 
\begin{lemma}\label{lemma:clip_trick}
For any $\tau>0$ and $g\in\mathbb{R}^d$, $\| \mathbf{clip}_\tau(g) - g\|  \leq \max( \| g \| - \tau ,0)$.	
\end{lemma}		
\begin{proof}
	From the definition of the clipping operator and the Euclidean norm, 
	\begin{align*}
		\| \mathbf{clip}_\tau(g) - g\| 
  & =  (\|  g \| - \tau ) \cdot \mathbbm{1}(\| g \| > \tau) + 0 \cdot \mathbbm{1}(\| g \| \leq \tau) \\
  & \leq \max( \| g \| - \tau,0). 
	\end{align*}
\end{proof}

Now, we bound $\| \eta^k\|^2$ in three steps: 

\paragraph*{\bf Step 1) Bound $\|\eta^k\|^2$ in terms of $\frac{1}{n}\sum_{i=1}^n\| \mathbf{clip}_\tau(g_k^i) - g_i^k\|^2$}
From Lemma~\ref{lemma:eta_k_supp} with $G(g_i^k) =\mathbf{clip}_\tau(g_k^i)$ and $\theta=2$, and by the fact that $\| g_i^k - \nabla f_i(x^k) \|^2 \leq \sigma^2$, 
\begin{align*}
	\| \eta^{k} \|^2 
 & \leq  \frac{2}{n}\sum_{i=1}^n \| \mathbf{clip}_\tau(g_k^i) - g_i^k\|^2 + \frac{2}{n}\sum_{i=1}^n \| g_i^k - \nabla f_i(x^k) \|^2 \\
 &\leq  \frac{2}{n}\sum_{i=1}^n \| \mathbf{clip}_\tau(g_k^i) - g_i^k\|^2 + 2\sigma^2.
\end{align*}
\paragraph*{\bf Step 2) Bound $\frac{1}{n}\sum_{i=1}^n \|\mathbf{clip}_\tau(g_k^i) - g_i^k\|^2$}
By Lemma~\ref{lemma:clip_trick}, and the fact that $\tau > 0$ and that $\| x\| \geq 0$ for $x\in\mathbb{R}^d$,
\begin{align*}
   \frac{1}{n}\sum_{i=1}^n\| \mathbf{clip}_\tau(g_k^i) - g_i^k\|^2 
   & \leq \max\left( \frac{1}{n}\sum_{i=1}^n[\| g_i^k \| - \tau]^2, 0 \right) \\ 
   & \leq \max \left( \frac{1}{n}\sum_{i=1}^n\| g_i^k \|^2 + \tau^2 , 0 \right).
\end{align*}
We bound $ \frac{1}{n}\sum_{i=1}^n\| \mathbf{clip}_\tau(g_k^i) - g_i^k\|^2 $ by bounding $\frac{1}{n}\sum_{i=1}^n\| g_i^k \|^2$.
Since, by \eqref{eqn:theta_x_y} with $\theta=2$, and by the fact that $f_i(x)$ has $L$-Lipschitz continuous gradient, that $\| g_i^k - \nabla f_i(x^k) \|^2 \leq \sigma^2$ and that $f(x^k)-f(x^\star) \leq \delta$,   
\begin{align*}
    \frac{1}{n}\sum_{i=1}^n\| g_i^k \|^2
    & \leq   \frac{2}{n}\sum_{i=1}^n \| g_i^k - \nabla f_i(x^k)\|^2 + \frac{2}{n}\sum_{i=1}^n \| \nabla f_i(x^k)\|^2  \\ 
    & \leq 2\sigma^2 + 4L [f(x^k) - f(x^\star)] \leq 2\sigma^2 + 4L \delta, 
\end{align*}
we have
\begin{align*}
   \frac{1}{n}\sum_{i=1}^n\| \mathbf{clip}_\tau(g_k^i) - g_i^k\|^2 
   & \leq \max(2\sigma^2 + 4L \delta + \tau^2, 0). 
\end{align*}
\paragraph*{\bf Step 3) Complete the upper-bound for $\| \eta^k\|^2$}
Plugging the upper-bound for $ \frac{1}{n}\sum_{i=1}^n\| \mathbf{clip}_\tau(g_k^i) - g_i^k\|^2$ into $\| \eta^k\|^2$ yields 
\begin{align*}
\| \eta^{k} \|^2 \leq 2\max(2\sigma^2 + 4L\delta + \tau^2, 0 ) + 2\sigma^2.  
\end{align*}
Finally taking the expectation, we complete the proof. 

\subsection{Proof of Proposition~\ref{prop:stochastic_composite}}\label{app:prop:stochastic_composite}
\noindent If each $g_{i,j}(x)$ is $\ell_g$-Lipschitz continuous and has $L_g$-Lipschitz continuous gradient, then by Cauchy-Schwartz's inequality $g_i(x) = \frac{1}{m_g}\sum_{j=1}^{m_g} g_{i,j}(x)$ is also $\ell_g$-Lipschitz continuous and has $L_g$-Lipschitz continuous gradient. Similarly, if each $F_{i,j}(x)$ is $\ell_F$-Lipschitz continuous and has $L_F$-Lipschitz continuous gradient, then by Cauchy-Schwartz's inequality $F_i(x) = \frac{1}{m_F}\sum_{j=1}^{m_F} F_{i,j}(x)$ is also $\ell_F$-Lipschitz continuous and has $L_F$-Lipschitz continuous gradient.

\subsubsection{Proof of Proposition~\ref{prop:stochastic_composite}-\ref{prop:stochastic_composite_1}}
\noindent We next prove the first statement. By the triangle inequality and by the fact that $\nabla f(x) = \frac{1}{n}\sum_{i=1}^n \nabla f_i(x)$ and  $\nabla f_i(x)= \langle \nabla g_i(x), \nabla F_i(g_i(x))\rangle$,
\begin{align*}
     \| \nabla f(x) - \nabla f(y) \| 
    & \leq  \frac{1}{n}\sum_{i=1}^n \| \nabla f_i(x) - \nabla f_i(y) \| \\
    & = \frac{1}{n} \sum_{i=1}^n \| T_1 + T_2 \| 
     \leq \frac{1}{n}\sum_{i=1}^n \left( \bar T_1 + \bar T_2 \right),
\end{align*}
where $T_1 = \langle \nabla g_i(x), \nabla F_i(g_i(x))\rangle - \langle \nabla g_i(y), \nabla F_i(g_i(x))\rangle$, $T_2= \langle \nabla g_i(y), \nabla F_i(g_i(x))\rangle  - \langle \nabla g_i(y), \nabla F_i(g_i(y))\rangle $, $\bar T_1 = \| \nabla F_i(g_i(x)) \|\cdot \| \nabla g_i(x) - \nabla g_i(y)  \|$, and $\bar T_2 = \| \nabla g_i(y) \| \cdot \| \nabla F_i(g_i(x)) - \nabla F_i(g_i(y)) \|$.

We bound $\| \nabla f(x) -\nabla f(y)\|$ by bounding $\bar T_1$ and $\bar T_2$.
First, by the $L_g$-Lipschitz continuity of $\nabla g_i(x)$ and the $\ell_F$-Lipschitz continuity of $F_i(x)$, we have 
\begin{align*}\bar T_1 \leq L_g \ell_F \|x-y\|.
\end{align*}
Second, by the $\ell_g$-Lipschitz continuity of $g_i(x)$ and the $L_F$-Lipschitz continuity of $\nabla F_i(x)$, we get 
\begin{align*}
\bar T_2 
& \leq \ell_g L_F \| g_i(x) - g_i(y) \| \leq \ell_g^2 L_F \| x - y \|.
\end{align*}
Finally, plugging the upper-bounds of $\bar T_1$ and $\bar T_2$, we obtain the upper-bound for $\| \nabla f(x)-\nabla f(y)\|$.

\subsubsection{Proof of Proposition~\ref{prop:stochastic_composite}-\ref{prop:stochastic_composite_2}} We finally prove the second statement. 

\noindent Let $\bar v^k := \frac{1}{\vert \mathcal{S}_g^k \vert} \sum_{j\in\mathcal{S}_g^k} g_{i,j}(x^k)$ and $\bar z^k:=\frac{1}{\vert \mathcal{S}_g^k \vert} \sum_{j\in\mathcal{S}_g^k} \nabla g_{i,j}(x^k)$. 
From the definition of the Euclidean norm, and by the fact that $\eta^k = \frac{1}{n}\sum_{i=1}^n \tilde \nabla f_i(x^k) - \nabla f_i(x^k)$, 
\begin{align*}
 \| \eta^k \|^2 
& \mathop{\leq}\limits^{\eqref{eqn:Jensen}} \frac{1}{n}\sum_{i=1}^n \| \tilde \nabla f_i(x^k) - \nabla f_i(x^k)\|^2, 
\end{align*}
where $\tilde \nabla f_i(x^k) = \left\langle \bar z^k ,  \frac{1}{\vert \mathcal{S}_F^k \vert} \sum_{j\in\mathcal{S}_F^k} \nabla F_{i,j} \left(  \bar v^k\right) \right\rangle$ and $\nabla f_i(x^k)=\left\langle  \nabla g_i(x^k) ,  \nabla F_i \left( g_i(x^k)\right) \right\rangle$. 
Next, by  the fact that $\|x+y+z \|^2 \leq 3\|x \|^2 + 3\|y\|^2+3\|z\|^2$ with $x= \left\langle \bar z^k ,  \frac{1}{\vert \mathcal{S}_F^k \vert} \sum_{j\in\mathcal{S}_F^k} \nabla F_{i,j} \left(  \bar v^k\right) \right\rangle -  \left\langle \bar z^k ,  \nabla F_i(\bar v^k)\right\rangle$, $y=\left\langle \bar z^k ,  \nabla F_i(\bar v^k)\right\rangle - \left\langle \nabla g_i(x^k) ,  \nabla F_i(\bar v^k)\right\rangle$, and $z= \left\langle \nabla g_i(x^k) ,  \nabla F_i(\bar v^k)\right\rangle- \left\langle  \nabla g_i(x^k) ,  \nabla F_i \left( g_i(x^k)\right) \right\rangle$, 
\begin{align}
    \|  \eta^k \|^2 
\leq  & 3 \| \bar z^k \|^2 T_1 + 3\|  \nabla F_i(\bar v^k) \|^2 T_2+ 3 \| \nabla g_i(x^k) \|^2 T_3, \label{eqn:eta_composite_bounds_notfinal}
\end{align}
where  $T_1= \left\| \frac{1}{\vert \mathcal{S}_F^k \vert} \sum_{j\in\mathcal{S}_F^k} \nabla F_{i,j} \left(  \bar v^k\right) -    \nabla F_i(\bar v^k) \right\|^2$, $T_2 = \left\|  \bar z^k -  \nabla g_i(x^k)  \right\|^2$, and $T_3= \left\|  \nabla F_i(\bar v^k) -   \nabla F_i \left( g_i(x^k)\right) \right\|^2$.
Since each $g_{i,j}(x)$ is $\ell_g$-Lipschitz continuous and each $F_{i,j}(x)$ is $\ell_F$-Lipschitz continuous, by \eqref{eqn:Jensen}  we have
\begin{align*}
    \| \bar z^k \|^2 & \leq \frac{1}{\vert \mathcal{S}_g^k \vert} \sum_{j\in\mathcal{S}_g^k} \| \nabla g_{i,j}(x^k)\|^2 \leq \ell_g^2, \\ 
    \| \nabla g(x^k) \|^2 & \leq \frac{1}{m_g}\sum_{j=1}^{m_g} \| \nabla g_{i,j}(x^k) \|^2 \leq \ell_g^2, \quad \text{and} \\
    \|  \nabla F_i \left( v^k \right) \|^2 & \leq \frac{1}{m_F}\sum_{j=1}^{m_F} \| \nabla F_{i,j}(v^k) \|^2 \leq \ell_F^2.
\end{align*}
Therefore, plugging these results into \eqref{eqn:eta_composite_bounds_notfinal},
\begin{eqnarray*}
    \|  \eta^k \|^2  &\leq&   3 \ell_g^2 T_1  + 3 \ell_F^2 T_2+ 3 \ell_g^2 T_3. 
\end{eqnarray*}
Since each $F_{i,j}(x)$ has $L_F$-Lipschitz continuous gradient, $F_i(x)$ has also $L_F$-Lipschitz continuous gradient. Hence, using this fact and taking the expectation, 
\begin{eqnarray*}
    \mathbf{E}\|  \eta^k \|^2  &\leq &   3 \ell_g^2 \mathbf{E} [T_1]  + 3 \ell_F^2 \mathbf{E}[T_2]  + 3 \ell_g^2 L_F^2 \mathbf{E}[\bar T_3],
\end{eqnarray*}
where $\bar T_3 = \| \bar v^k - g_i(x^k)\|^2$.
Assuming that $\mathcal{S}_g^k$ and $\mathcal{S}_F^k$ are sampled uniformly at random, we get $\mathbf{E}\left[ \frac{1}{\vert \mathcal{S}_g^k \vert} \sum_{j\in\mathcal{S}_g^k} g_{i,j}(x^k) \right] = g_i(x^k)$, $\mathbf{E}\left[ \frac{1}{\vert \mathcal{S}_g^k \vert} \sum_{j\in\mathcal{S}_g^k} \nabla g_{i,j}(x^k) \right] = \nabla g_i(x^k)$ and $\mathbf{E}\left[ \frac{1}{\vert \mathcal{S}_F^k \vert} \sum_{j\in\mathcal{S}_F^k} \nabla F_{i,j} \left(  \bar v^k\right) \right] = \nabla F_i(\bar v^k)$. In addition, $\mathbf{E}\left\| \frac{1}{\vert \mathcal{S}_F^k \vert} \sum_{j\in\mathcal{S}_F^k} \nabla F_{i,j} \left(  \bar v^k\right) -    \nabla F_i(\bar v^k) \right\|^2  \leq  \frac{\sigma_F^2}{\vert \mathcal{S}_F^k \vert}$, $ \mathbf{E}\left\|  \bar z^k -  \nabla g_i(x^k)  \right\|^2  \leq \frac{\sigma_{\nabla g}^2}{\vert  \mathcal{S}_g^k \vert}$, and $\mathbf{E}\left\|  \bar v^k -    g_i(x^k) \right\|^2  \leq \frac{\sigma_g^2}{\vert  \mathcal{S}_g^k \vert}$.
Therefore, 
\begin{eqnarray*}
    \mathbf{E}\|  \eta^k \|^2  \leq   3 \ell_g^2 \frac{\sigma_F^2}{\vert \mathcal{S}_F^k \vert}   + 3 \ell_F^2 \frac{\sigma_{\nabla g}^2}{\vert  \mathcal{S}_g^k \vert}   + 3 \ell_g^2 L_F^2 \frac{\sigma_g^2}{\vert  \mathcal{S}_g^k \vert}.
\end{eqnarray*}
Finally, setting $\vert \mathcal{S}_F^k \vert = S_F$ and $\vert  \mathcal{S}_g^k \vert= S_g$ yields the  result.

\subsection{Proof of Proposition~\ref{prop:MAML_SC}}\label{app:prop:MAML_SC}
\noindent Let $\ell_{i}(x,a^j_i,b^j_i)$ be $\ell_l$-Lipschitz continuous and have $L_l$-Lipschitz continuous gradient. 
By the fact that $F_{i,j}(x)=\ell_{i}(x,a^j_i,b^j_i)$, $ \| F_{i,j}(x) - F_{i,j}(y) \| \leq \ell_l \| x-y \|,$ and $ \| \nabla F_{i,j}(x) - \nabla F_{i,j}(y) \| 
    \leq L_l \| x-y \|$.
Next, by the fact that $g_{i,j}(x)=x-\gamma \nabla \ell_{i}(x,a^j_i,b^j_i)$, 
\begin{align*}
    \| g_{i,j}(x) - g_{i,j}(y) \| & \leq \|x-y\| + \gamma T_1   \leq (1+\gamma L_l) \|x-y\|,
\end{align*}
where $T_1 =  \| \nabla \ell_{i}(x,a^j_i,b^j_i) - \nabla \ell_{i}(y,a^j_i,b^j_i) \|$. 
In addition, we can show that $\nabla g_{i,j}(x)=I - \gamma \nabla^2  \ell_{i}(x,a^j_i,b^j_i)$, and that
\begin{align*}
    \| \nabla g_{i,j}(x) - \nabla g_{i,j}(y)  \| 
    & \leq \gamma \|  \nabla^2 \ell_{i}(x,a^j_i,b^j_i) - \nabla^2 \ell_{i}(y,a^j_i,b^j_i) \| \\ 
    & \leq \gamma(B(x) +B(y)) \leq 2\gamma L_l,
\end{align*}
where $B(x) = \| \nabla^2 \ell_{i}(x,a^j_i,b^j_i) \|$.

\section{Conclusion}\label{sec:conclusion}
\noindent We unify an analysis framework for  parallel stochastic momentum methods with biased gradients. 
We show that biased momentum methods attain the worst-case bound similar to biased SGD for general non-convex and $\mu$-PL problems. 
We apply our results to parallel momentum methods with compressed, clipped, and composite gradients, including distributed MAML. 
Numerical experiments demonstrated that biased momentum methods outperform biased gradient descent, achieving faster convergence and higher accuracy.
As future work, establishing worst-case bounds based on our analysis framework in other applications, including reinforcement learning and risk-aware learning, would be considered a worthwhile study.

\appendices
\section{Basic Facts}
\noindent We use the following facts from linear algebra: for any $x,y\in\mathbb{R}^d$ and $\theta>0$,
\begin{align}
	2 \langle x, y \rangle & = \|  x\|^2 + |y\|^2 - \| x-y \|^2. \label{eqn:product_eq}  \\
	2 \langle x, y \rangle &\leq  \theta\|  x\|^2 + \theta^{-1}\|y\|^2. \label{eqn:theta_product}  \\
	\|  x + y\|^2 &\leq (1+\theta)\| x \|^2 + (1+\theta^{-1})\|y\|^2. \label{eqn:theta_x_y}
\end{align}	 
For vectors $x_1,x_2,\ldots,x_N \in\mathbb{R}^d$, Jensen's inequality and the convexity of the squared norm yields 
\begin{align}\label{eqn:Jensen}
	\left\Vert  \frac{1}{N}\sum_{i=1}^N x_i \right\Vert^2 \leq \frac{1}{N}\sum_{i=1}^N \| x_i\|^2.
\end{align}	 
The next lemma allows us to obtain the upper bound for the step-size satisfying the specific inequality. 
\begin{lemma}[Lemma 5 of \cite{richtarik2021ef21}]\label{lemma:stepsize_range}
Let $a,b,c>0$. If $0 < \gamma \leq (\sqrt{ \nicefrac{c}{a} } +\nicefrac{b}{a})^{-1}$, then $\frac{a}{\gamma}-b-c\gamma \geq 0$.
\end{lemma}
		
Furthermore, we introduce the next lemma, which is useful for deriving the upper bound for $\| \eta^{k} \|^2$ in Section~\ref{sec:App}.
\begin{lemma}\label{lemma:eta_k_supp}
Let $f(x)=(1/n)\sum_{i=1}^n f_i(x)$ and $\eta^{} = (1/n)\sum_{i=1}^n G(g_i^k) - \nabla f(x^k)$ for any operator $G:\mathbb{R}^d\rightarrow\mathbb{R}^d$. Then, for $\theta>0$,
\begin{align}
	\| \eta^{k} \|^2 
 \leq   (1+\theta)  T_1^k + (1+1/\theta) T_2^k, \label{eqn:eta_k}
\end{align}	
where $T_1^k = \frac{1}{n}\sum_{i=1}^n \| G(g_i^k) - g_i^k\|^2$ and $T_2^k = \frac{1}{n}\sum_{i=1}^n \| g_i^k - \nabla f_i(x^k) \|^2$.  
\end{lemma}	
\begin{proof}
By the fact that $\nabla f(x)=(1/n)\sum_{i=1}^n \nabla f_i(x)$, and from the definition of the Euclidean norm and $\eta^{k}$, we have $\| \eta^{k} \|^2 = \left\|  ({1}/{n})\sum_{i=1}^n [G(g_i^k) - \nabla f_i(x^k)] \right\|^2.$
Finally, by  \eqref{eqn:Jensen} and then by \eqref{eqn:theta_x_y}, we complete the proof. 
\end{proof}	

\section{Additional Numerical Evaluations}
\label{AdditionalCurves}
\noindent More results for both the MNIST dataset and the FashionMNIST dataset are included in the following link: \url{https://github.com/AliBeikmohammadi/DistributedSGDM/} (See Figures 2-9), utilizing FCNN and ResNet-18 model, respectively. 

\balance
\bibliographystyle{IEEEtran}
\bibliography{sample_revised.bib}
\begin{IEEEbiography}[{\includegraphics[width=1in,height=1.25in,clip,keepaspectratio]{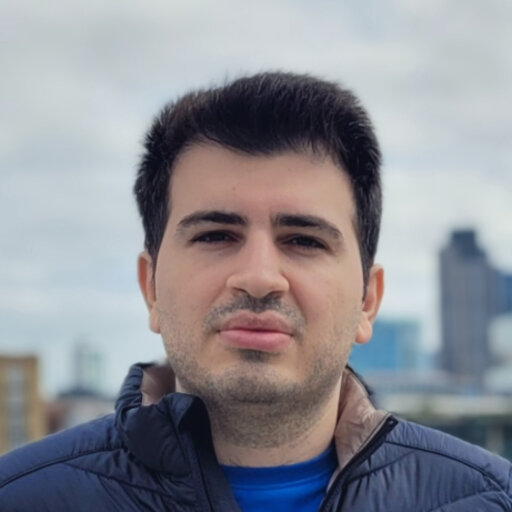}}]{Ali Beikmohammadi}
earned his B.Sc. in Electrical Engineering from Bu-Ali Sina University, Hamedan, Iran, in 2017, and subsequently completed his M.Sc. in Electrical Engineering at Amirkabir University of Technology, Tehran, Iran, in 2019. Currently, he is a Ph.D. candidate in the Department of Computer and Systems Sciences at Stockholm University, Sweden, focusing on research areas such as Reinforcement Learning, Deep Learning, and Federated Learning, both in theory and applications.
\end{IEEEbiography}
\begin{IEEEbiography}[{\includegraphics[width=1in,height=1.25in,clip,keepaspectratio]{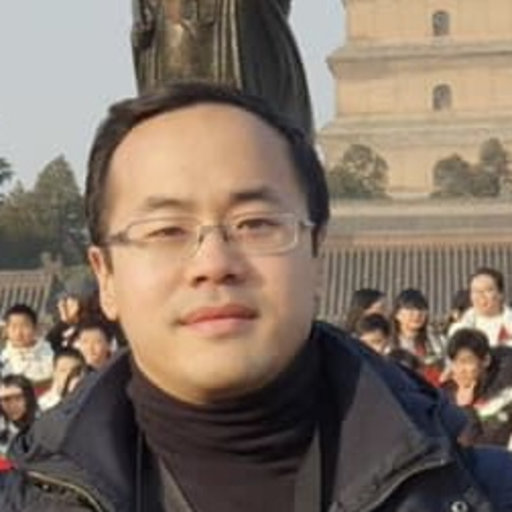}}]{Sarit Khirirat}
 received his B.Eng. in Electrical Engineering from Chulalongkorn University, Thailand, in 2013; the Master’s degree in Systems, Control and Robotics from KTH Royal Institute of Technology, Sweden, in 2017; and the PhD at the Division of Decision and Control Systems from the same institution in 2022, supported by the Wallenberg AI, Autonomous Systems and Software program, Sweden’s largest individual research funding program. He is currently a postdoctoral fellow at King Abdullah University of Science and Technology (KAUST). His research interests include distributed optimization algorithms for federated learning applications. 
\end{IEEEbiography}
\begin{IEEEbiography}[{\includegraphics[width=1in,height=1.25in,clip,keepaspectratio]{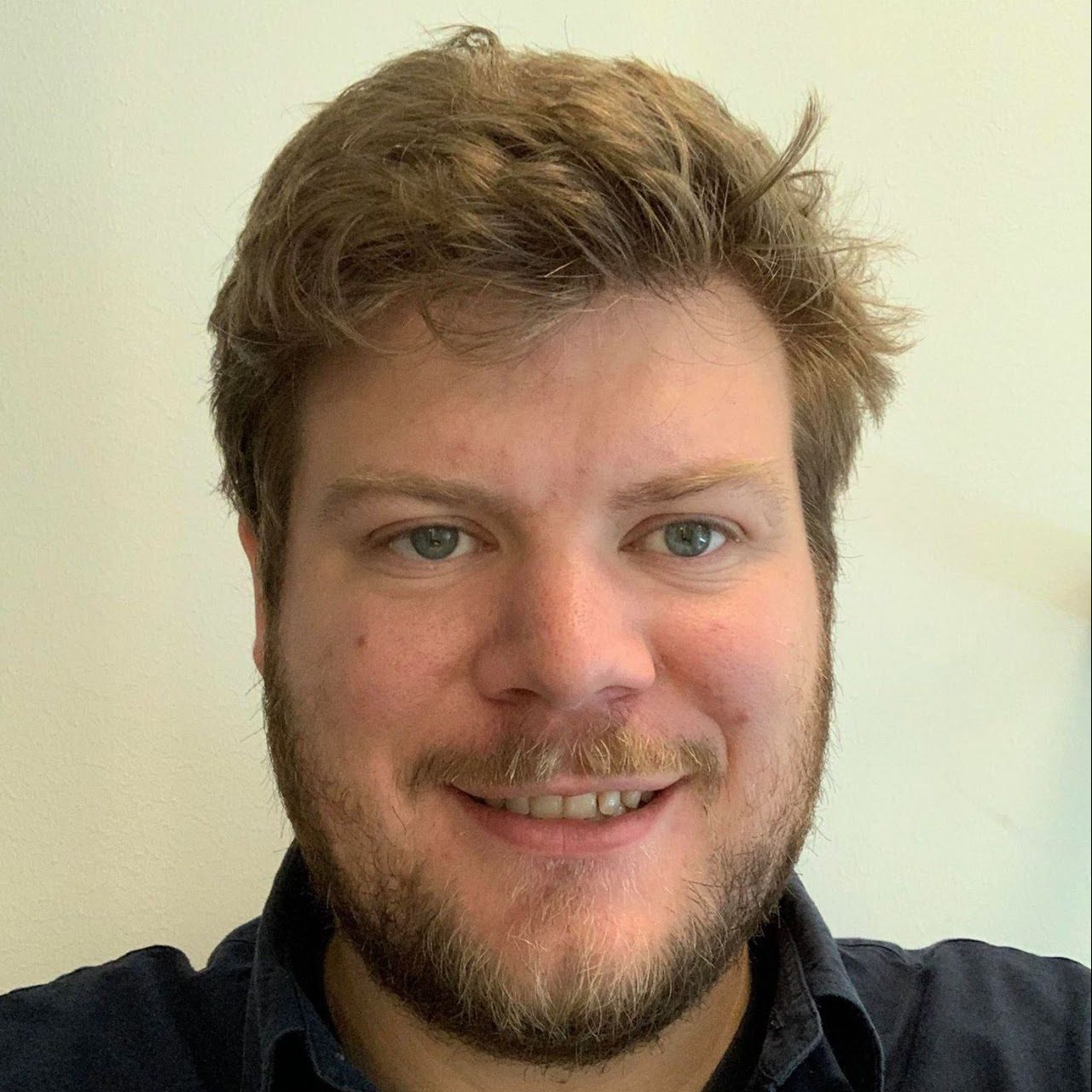}}]{Sindri Magn\'usson}
is an Associate Professor in the Department of Computer and Systems Science at Stockholm University, Sweden. He received a B.Sc. degree in Mathematics from the University of Iceland, Reykjav´ık Iceland, in 2011, a Masters degree in Applied Mathematics (Optimization and Systems Theory) from KTH Royal Institute of Technology, Stockholm, Sweden, in 2013, and the PhD in Electrical Engineering from the same institution, in 2017. He was a postdoctoral researcher 2018-2019 at Harvard University, Cambridge, MA, and a visiting PhD student at Harvard University for 9 months in 2015 and 2016. His research interests include large-scale distributed/parallel optimization, machine learning, and control.
\end{IEEEbiography}

\end{document}